\documentclass[times,twocolumn,preprint]{elsarticle}

\usepackage{medima}
\usepackage{framed,multirow}

\usepackage{subcaption} % used for subfigure

\usepackage{comment} % for comment
\usepackage{booktabs}

\usepackage{pgfplots}
\usepackage{pgfplotstable}

\usepackage{amsmath}
\usepackage{amssymb}
\usepackage{bbm}

\usepackage{physics}
\usepackage{rotating}

\usepackage{hyperref}

\newcommand{\ufdm}{u^{FDM}}
\newcommand{\bufdm}{\bar{u}^{FDM}}
\newcommand{\upinn}{u^{PINN}}
\newcommand{\ugt}{u^{GT}}
\newcommand{\busph}{\bar{u}^{sph}}

\newcommand{\segf}{y^{\rm FLAIR}}
\newcommand{\segt}{y^{\rm T1Gd}}
\newcommand{\ucf}{u_c^{\rm FLAIR}}
\newcommand{\uct}{u_c^{\rm T1Gd}}

\newcommand{\bucf}{{\bar u}_c^{\rm FLAIR}}
\newcommand{\buct}{{\bar u}_c^{\rm T1Gd}}

\newcommand{\segs}{y^{\rm s}}
\newcommand{\ucs}{u_c^{\rm s}}

\newcommand{\rf}{R^{\rm FLAIR}}
\newcommand{\rt}{R^{\rm T1Gd}}

\newcommand{\segfet}{y^{\rm FET}}

\newcommand{\corfet}{corr\textsuperscript{FET}}

\newcommand{\bu}{\bar{u}}
\newcommand{\ra}{\bD/\brho}
\newcommand{\bD}{\bar{D}}
\newcommand{\bT}{\bar{T}}
\newcommand{\bx}{\bar{x}}
\newcommand{\bt}{\bar{t}}
\newcommand{\brho}{\bar{\rho}}
\newcommand{\bL}{\bar{L}}

\newcommand{\cR}{\mathcal{R}}
\newcommand{\cD}{\mathcal{D}}
\newcommand{\muR}{\mu_\mathcal{R}}
\newcommand{\muD}{\mu_\mathcal{D}}
\newcommand{\tend}{t_{\rm end}}

\newcommand{\vbx}{{\vb x}}
\newcommand{\icx}{{\vb x_0}}

\newcommand{\xdat}{{\vb x}^{\rm dat}}
\newcommand{\xres}{{\vb x}^{\rm r}}

\newcommand{\Nres}{N_{\rm r}}
\newcommand{\Ndat}{N_{\rm dat}}

\newcommand{\wseg}{w_{\rm SEG}}
\newcommand{\wfet}{w_{\rm FET}}

\newcommand{\ltot}{\mathcal{L}_{\rm total}}
\newcommand{\lres}{\mathcal{L}_{\rm PDE}}
\newcommand{\ldat}{\mathcal{L}_{\rm data}}

\newcommand{\lseg}{\mathcal{L}_{\rm SEG}}
\newcommand{\lfet}{\mathcal{L}_{\rm FET}}
\newcommand{\lchar}{\mathcal{L}_{\rm char}}

\newcommand{\ctvrtog}{\rm CTV^{\rm RTOG}}
\newcommand{\ctvp}{\rm CTV^{\rm P}} %personalized CTV

\newcommand{\petseg}{\textbf{FET+SEG}}
\newcommand{\segonly}{\textbf{SEG}}
\newcommand{\utall}{\textbf{u-t-all}}
\newcommand{\utend}{\textbf{u-t-end}}
\newcommand{\jana}{Lipková et al. (2019)}

\begin{document}

\begin{frontmatter}

\title{Personalized Predictions of Glioblastoma Infiltration: Mathematical Models, Physics-Informed Neural Networks and Multimodal Scans}

%\author[1]{TBD}
%\address[1]{Planet Earth}

 \author[1]{Ray Zirui \snm{Zhang}\corref{cor1}}
 \cortext[cor1]{Corresponding authors: zirui.zhang@uci.edu, jlowengr@uci.edu}
 \author[4]{Ivan \snm{Ezhov}}
 \author[3]{Michał \snm{Balcerak}}
 \author[2]{Andy \snm{Zhu}}
 \author[4]{Benedikt \snm{Wiestler}}
 \author[3]{Bjoern \snm{Menze}}
 \author[1,5]{John S. \snm{Lowengrub}\corref{cor1}}
  %\cortext[cor1]{Co-corresponding author: jlowengr@uci.edu}

 \address[1]{Department of Mathematics, University of California Irvine, USA}
 \address[4]{Technical University of Munich, Germany}
 \address[2]{Carnegie Mellon University, USA}
 \address[3]{University of Zurich, Switzerland}
 \address[5]{Department of Biomedical Engineering, University of California Irvine, USA}

\begin{abstract}
Predicting the infiltration of Glioblastoma (GBM) from medical MRI scans is crucial for understanding tumor growth dynamics and designing personalized radiotherapy treatment plans.
Mathematical models of GBM growth can complement the data in the prediction of spatial distributions of tumor cells.
However, this requires estimating patient-specific parameters of the model from clinical data, which is a challenging inverse problem due to limited temporal data and the limited time between imaging and diagnosis.
This work proposes a method that uses Physics-Informed Neural Networks (PINNs) to estimate patient-specific parameters of a reaction-diffusion partial differential equation (PDE) model of GBM growth from a single 3D structural MRI snapshot.
PINNs embed both the data and the PDE into a loss function, thus integrating theory and data.
Key innovations include the identification and estimation of characteristic non-dimensional parameters, a pre-training step that utilizes the non-dimensional parameters and a fine-tuning step to determine the patient specific parameters.
Additionally, the diffuse domain method is employed to handle the complex brain geometry within the PINN framework.
The method is validated both on synthetic and patient datasets, and shows promise for real-time parametric inference in the clinical setting for personalized GBM treatment.
\end{abstract}

% \maketitle 


\end{frontmatter}

\section{Introduction}\label{s:intro}

% what is the problem
Glioblastoma (GBM) is a common and very aggressive form of brain tumor, with a median survival of 15 months \citep{stuppHighgradeGliomaESMO2014,stuppRadiotherapyConcomitantAdjuvant2005}.
The current standard of care for GBM includes immediate resection of the visible tumor after diagnosis, followed by radiotherapy and chemotherapy targeting residual tumor \citep{fernandesCurrentStandardsCare2017}. 
Despite intensive treatment, the prognosis of GBM remains poor, as almost all GBMs recur.
The high recurrence rate of GBM is attributed to its infiltrative nature. 
Instead of forming a clear boundary, the tumor invades into the surrounding normal-looking brain tissue \citep{souhamiRandomizedComparisonStereotactic2004,halperinRadiationTherapyTreatment1989}
The infiltration pattern of tumor cells is heterogeneous and varies significantly among patients.
Consequently, the standard clinical target volume (CTV) of radiotherapy, which extends 1.5-3 cm uniformly from the visible tumor, may be insufficient for optimal coverage and distribution \citep{stuppRadiotherapyConcomitantAdjuvant2005}.
A critical step towards enhancing radiotherapy treatment planning requires the estimation of tumor cell distributions in a data-driven and patient-specific manner
\citep{lipkovaPersonalizedRadiotherapyDesign2019,unkelbachRadiotherapyPlanningGlioblastoma2014,rocknePredictingEfficacyRadiotherapy2010,rocknePatientspecificComputationalModel2015a,rockne2019MathematicalOncology2019}.

% what are some modeling approaches
Mathematical models of tumor growth are instrumental in predicting tumor progression and guiding individualized treatment decisions for patients \citep{baldockPatientSpecificMathematicalNeuroOncology2013,lipkovaPersonalizedRadiotherapyDesign2019,butnerMathematicalModelingCancer2022,ezhovLearnMorphInferNewWay2023,chaudhuriPredictiveDigitalTwin2023,subramanianEnsembleInversionBrain2023,lorenzoPatientspecificMechanisticModels2023,wuIntegratingMechanismbasedModeling2022}.
The growth of GBM is often modeled using partial differential equations (PDEs) of reaction-diffusion type, which describe the spatio-temporal evolution of tumor cell density and captures the main tumor behavior: proliferation and infiltration \citep{harpoldEvolutionMathematicalModeling2007}.
More advanced models incorporate additional features including the mass effect \citep{hogeaRobustFrameworkSoft2007,subramanianSimulationGlioblastomaGrowth2019,lipkovaModellingGliomaProgression2022}, angiogenesis \citep{macklinMultiscaleModellingNonlinear2009,sautMultilayerGroworGoModel2014,yan3DMathematicalModeling2017}, chemotaxis \citep{luNonlinearSimulationVascular2022,luBifurcationAnalysisFree2023}, and multiple species \citep{curtinSpeedSwitchGlioblastoma2020,yan3DMathematicalModeling2017}. 
For comprehensive reviews on mathematical modeling of GBM, see \citet{alfonsoBiologyMathematicalModelling2017,falcoSilicoMathematicalModelling2021,jorgensenDatadrivenSpatiotemporalModelling2023}.

% we work on the inverse problem, what are the method and challenges
To apply these models in a clinical setting, it is imperative to estimate patient-specific parameters of the PDEs from clinical data, which is a challenging inverse problem. Besides being useful in solving the PDE to predict tumor cell density profiles, these patient-specific parameters are valuable on their own as biophysics-based features to improve patient stratification and survival prediction. For example, recent studies show that these biophysical features correlate better with survival time than volumetric features from images \citep{subramanianEnsembleInversionBrain2023}. 
Further, the ratio of proliferation rate to diffusion rate is key in evaluating the effectiveness of supramarginal resection for IDH-wildtype GBM patients \citep{tripathiIDHWildtypeGlioblastoma2021}.

The inverse problem is commonly solved in the framework of PDE constrained optimization \citep{mangBiophysicalModelingBrain2012,hogeaImagedrivenParameterEstimation2008,scheufeleImageDrivenBiophysicalTumor2020,scheufeleFullyAutomaticCalibration2021,subramanianWhereDidTumor2020,subramanianMultiatlasCalibrationBiophysical2020,subramanianEnsembleInversionBrain2023} or Bayesian inference \citep{menzeGenerativeApproachImageBased2011,lipkovaPersonalizedRadiotherapyDesign2019,ezhovNeuralParametersEstimation2019,ezhovGeometryAwareNeuralSolver2022}. 
Machine learning approaches have also been explored to speed up the computations,
including learning surrogate models for forward PDE systems \citep{ezhovGeometryAwareNeuralSolver2022}, 
learning mappings from observations to parameters directly \citep{patiEstimatingGlioblastomaBiophysical2021,ezhovLearnMorphInferNewWay2023,martensDeepLearningReactionDiffusion2022}, 
using dynamic mode decomposition to construct low-dimensional representations \citep{viguerieDataDrivenSimulationFisher2022},
and optimizing discrete loss frameworks that integrate the PDE and the data using mesh-based discretizations \citep{karnakovOptimizingDIscreteLoss2022,Balcerak2023}.
Instead of the full reaction-diffusion model, its asymptotic properties can be exploited to obtain an approximate anisotropic Eikonal approximation, which simplifies parameter estimation and tumor cell density prediction, albeit with a less detailed model \citep{konukogluExtrapolatingGliomaInvasion2010,konukogluImageGuidedPersonalization2010,unkelbachRadiotherapyPlanningGlioblastoma2014}.
% In a realistic clinical scenario, significant challenge arises as the temporal data is typically limited to a single-time snapshot, rendering the inverse problem ill-posed since multiple parameters can yield the same solution.
% Additionally, the limited time between imaging and diagnosis necessitate rapid inference 
% \citep{subramanianEnsembleInversionBrain2023,lipkovaPersonalizedRadiotherapyDesign2019}. 

% Different methods have been proposed to handle this challenge.
% In \citet{subramanianEnsembleInversionBrain2023}, the ill-posedness is handled by using multiple atlas and regularizing the initial condition.
% In the Bayesian approach \citep{lipkovaPersonalizedRadiotherapyDesign2019}, the joint distribution shows a strong correlation between the parameters, while individual parameters will have large variance.
% In \citep{ezhovLearnMorphInferNewWay2023,ezhovForloopAllYou2022}, the time-independent parameters are considered.
% Another challenge is the time complexity of 
% In addition, the time between imaging and diagnosis is usually limited, so the inference needs to be performed rapidly \citep{ezhovLearnMorphInferNewWay2023,ezhovForloopAllYou2022}. 

% why we use PINN
Physics-Informed Neural Networks (PINNs) have emerged as a novel method for solving inverse problems in a PDE constrained optimization framework \citep{karniadakisPhysicsinformedMachineLearning2021,raissiPhysicsinformedNeuralNetworks2019,jagtapPhysicsinformedNeuralNetworks2022,jagtapDeepLearningInverse2022,chenPhysicsinformedNeuralNetworks2020,zhengHomPINNsHomotopyPhysicsinformed2023,huangHomPINNsHomotopyPhysicsinformed2022a}.
In this approach, PINNs represent the PDE solution as a neural network and embed both the data and the PDE, via a mesh-free approach using automatic differentiation, into the loss function.
By minimizing the total loss, PINNs effectively solve the PDE and fit the data simultaneously, showcasing integration of mathematical models with data-driven learning processes. 
%In a related approach, the Optimizing Discrete Loss (ODIL) method also combines data and PDE into a loss function but uses a mesh-based discretization of the PDE rather than a neural network \citep{karnakovOptimizingDIscreteLoss2022}, and is recently used successfully in \citet{Balcerak2023} for glioma radiotherapy design.
This is particularly beneficial in the realm of medical image analysis, where clinical measurements are often sparse and expensive.
PINNs have been employed to combine the transcranial doppler data of blood flow velocity measurements and PDEs that govern the flow rate, pressure, and vessel area to estimate the brain hemodynamics \citep{sarabianPhysicsInformedNeuralNetworks2022}.
PINNs have also been used to calibrate the parameters in tracer-kinetic models with perfusion MR data for quantification of myocardial blood flow \citep{vanhertenPhysicsinformedNeuralNetworks2022}.
The potential for solving GBM inverse problems with PINNs was explored in \cite{zhuAcceleratingParameterInference2022} using synthetic cell density data.

% summary
% In this work, we employ PINNs to estimate the parameters of a reaction-diffusion PDE model for GBM from a single 3D structural MRI snapshot. 

\begin{figure*}[!t]
  \centering
  \includegraphics[keepaspectratio,width=\textwidth]{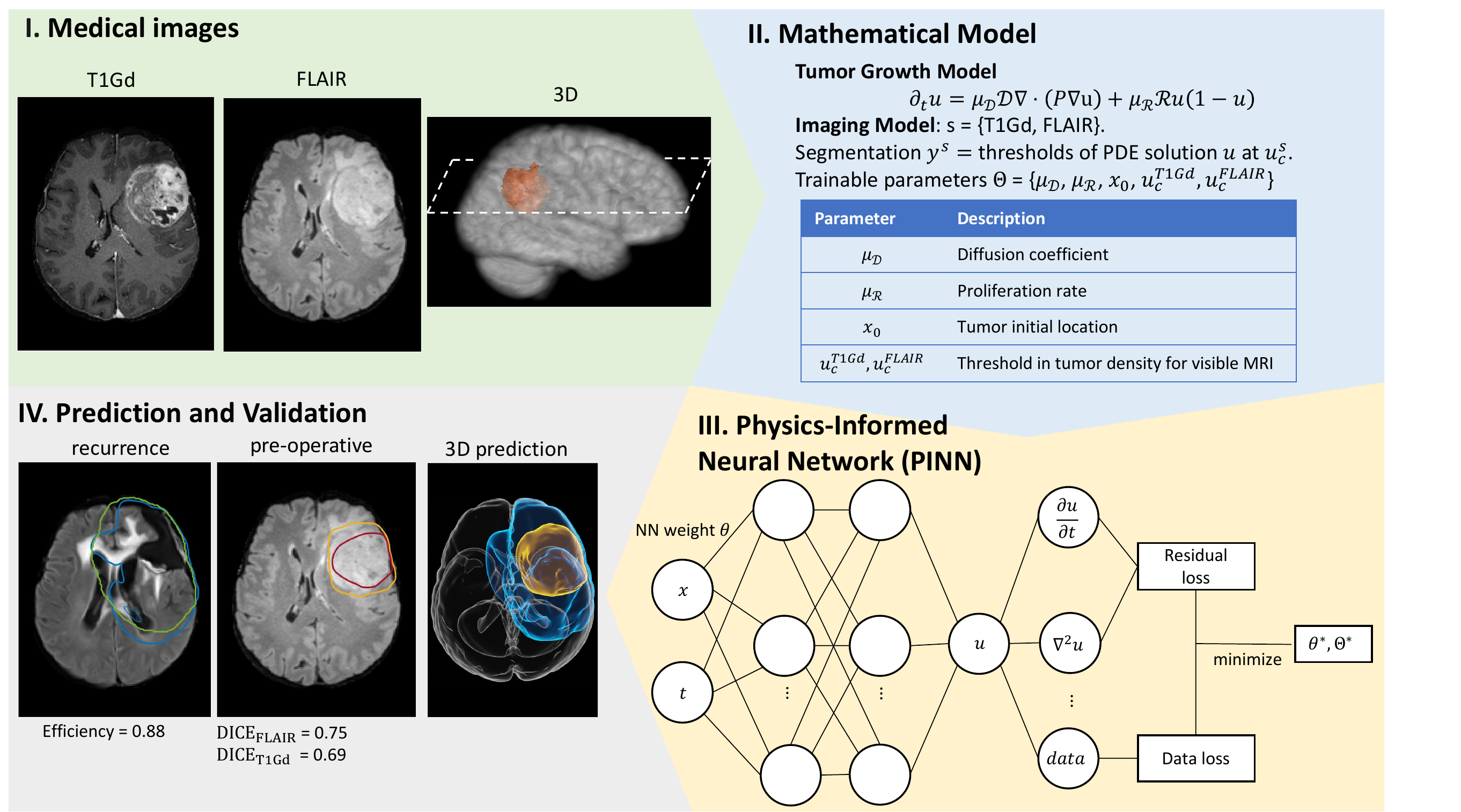}
  \caption{
    Overview of the parameter estimation framework.
    \textbf{I) Medical Images} show preoperative patient T1Gd and FLAIR scans.
    \textbf{II) Mathematical Model} includes a Partial Differential Equation (PDE) for the tumor cell density $u$, and an imaging model that relates medical images with $u$ by thresholding (segmentation). The unknown parameters that need to be estimated are tabulated.
    \textbf{III) Physics-informed Neural Network (PINN)} is used to solve the PDE and estimate parameters from data. See Fig. \ref{f:workflow} below for a detailed workflow.
    \textbf{IV) Prediction and Validation:} The estimated parameters can be used in the PDE model to predict the 3D tumor cell density (right), and can be validated via DICE scores (center).
    Left: The personalized clinical target volume (CTV, blue) based on model predictions is compared with Radiation Therapy Oncology Group (RTOG) CTV (green) to evaluate radiation volume and coverage of tumor recurrence (efficiency). 
         }
  \label{f:overview}
\end{figure*}

In this work, we make personalized predictions of GBM tumor cell density distributions beyond the mass visible in medical scans by using solutions of a reaction-diffusion PDE model with parameters estimated from 3D structural MRI snapshots at a single time point.
Fig.~\ref{f:overview} provides an overview of our approach. 
% overview fig part 1
Our data comprises a single snapshot of T1 gadolinium enhanced (T1Gd) and fluid attenuation inversion recovery (FLAIR) MRI scans.
% overview fig part 2
We combine the PDE model with an image model that relates the segmentation with the tumor cell density through thresholding. We overcome the challenge of inferring parameters using single time point data by identifying time-independent parameters, appropriately non-dimensionalizing the model and inferring the non-dimensional parameters.
% overview fig part 3
We use PINNs to solve the PDE and estimate the parameters from the data 
%(detailed in Section~\ref{s:method} and Fig.~\ref{f:workflow}).
% overview fig part 4
The estimated parameters are used to predict tumor cell densities and aid personalized radiotherapy treatment planning.
We validate our approach on synthetic data and patient data, where we can compute quantitative metrics such as DICE score of the segmentation, and compare the predicted infiltration pattern with observed tumor recurrence.
%The patient data with recurrence scans is taken from \jana{}.

% what are the contributions of this work
Our contributions are as follows:
(1) We show that PINN is an effective and efficient method for solving the inverse problem of PDE parameter estimation in GBM using structural MRI data at a single time point, which is the most common clinical data available.
(2) A simple procedure is introduced to estimate patient-specific non-dimensional characteristic parameters, which aid in scaling the PDE to address the single-time inference limitation.
(3) We propose an effective two-step training procedure: 
in the pre-training step, we leverage the patient-specific non-dimensional characteristic parameters to train the PINN to learn the characteristic solution of the PDE. 
This serves as an excellent starting point for the fine-tuning stage, where we train the PINN to learn patient-specific parameters.
(4) We introduce the diffuse domain method to handle the complex brain geometry in the PINN framework.

In the remainder of the paper, Section~\ref{s:method} elucidates the components of our methods, including 
the mathematical model of tumor growth (Section~\ref{ss:model}),
the workflow of the parameter estimation process (Section~\ref{ss:workflow}),
and the loss functions used for training the PINN (Section~\ref{ss:loss}).
The results are presented in Section~\ref{s:results}, where we first validate our approach on synthetic data (Section~\ref{ss:synthetic}), and then apply our approach to patient data (Section~\ref{ss:patient}).
In Section~\ref{ss:discussion},we also discuss the merits and limitations of our approach, compared with other methods,
Conclusions are presented in Section~\ref{s:conclusion}.
Additional technical details and results are given in the appendices.

\section{Methods}\label{s:method}

\subsection{Model} \label{ss:model}

\subsubsection{Tumor Model} 
\label{sss:pdemodel}
We consider the Fisher-KPP PDE, a fundamental model that describes the spatio-temporal evolution of the normalized tumor cell density $u(\vb x, t)$ in a 3D domain $\Omega$ consisting of white matter (WM) and grey matter (GM) regions \citep{harpoldEvolutionMathematicalModeling2007,swansonQuantitativeModelDifferential2000}. The PDE, with Neumann boundary conditions, is given by
\begin{equation} \label{eq:pde1}
    \begin{cases}
      \pdv{u}{t} = \grad \cdot ( \mathbb{D}(\vb x) \grad u) + \rho u (1-u) \quad \text{in }  \Omega \\
      \grad u \cdot \vb n = 0 \quad \text{on } \partial \Omega 
    \end{cases}
\end{equation}
where $\rho$ [1/day] is the proliferation rate and $\mathbb{D}(\vb x)$ [mm$^2$/day] is the diffusion tensor, and $\vb n$ is the outward normal vector on the boundary $\partial \Omega$.
We further assume that the diffusion is isotropic, and is a weighted sum of the diffusion coefficient in WM and GM,
$\mathbb{D}(\vb x) = D_w P_w(\vb x) +  D_g P_g(\vb x)$, 
where $P_w$, $P_g$ are the percentages of white and grey matter,
and $D_w$, $D_g$ [mm$^2$/day] are the diffusion coefficient in white and grey matter,
and $D_w = 10 D_g$ \citep{menzeGenerativeApproachImageBased2011,lipkovaPersonalizedRadiotherapyDesign2019,swansonQuantitativeModelDifferential2000}.
Therefore, the original PDE \eqref{eq:pde1} can be written as 
\begin{equation} \label{eq:pde}
    \pdv{u}{t} = D \grad \cdot (P(\vb x) \grad u) + \rho u (1-u) \quad \text{in }  \Omega \\
\end{equation}
where $D = D_w$ is the scalar diffusion coefficient in white matter, and $P(\vb x) = P_w(\vb x) + 10 P_g(\vb x)$ defines the geometry of the brain. We use an atlas-guided approach to obtain the brain geometry and tissue percentages. In particular, we register the brain atlas used in \cite{lipkovaPersonalizedRadiotherapyDesign2019} to patient T1w scans using rigid registration following \cite{lipkovaPersonalizedRadiotherapyDesign2019} . 
While rigid registration works well in many cases, some errors may arise 
 near the ventricles, especially when there is significant deformation of the brain due to the tumor.
It remains an active area of research to design registration methods that robustly account for significant distortions due to tumor mass effect and post-surgical resection \citep{lipkovaPersonalizedRadiotherapyDesign2019}. While we used a single brain atlas here, we could have used multiple atlases and chosen the atlas for registration that most closely matched the patient scans \citep{subramanianEnsembleInversionBrain2023}.
%
% which can be obtained by registering the atlas to the patient T1Gd scan using a rigid transformation \citep{lipkovaPersonalizedRadiotherapyDesign2019}.
We assume the initial condition is $u_0(\vb x) = 0.1 \exp(-0.1 |\vb x - \icx|^2)$, where $\icx$ denotes the initial location of the tumor.

%%%%%%%%%%%%%%%%%%%%%%
\subsubsection{Diffuse Domain Method} \label{sss:ddm}

A common way of enforcing the boundary condition using PINN requires sampling collocation points on the exact location of the boundary.
However, the brain geometry is complex and the image resolution is limited, therefore an exact parameterization and the normal vector field of the boundary is not easily obtainable.
In our work, the boundary condition is handled by the diffuse domain method \citep{lervagAnalysisDiffusedomainMethod2015,liSolvingPdesComplex2009}, which embeds the complex brain region $\Omega$ into a cubic box $\Omega_B$ by using a phase field function $\phi$ that replaces the sharp boundary of the domain with a smooth transition layer, and the PDE is modified:
\begin{equation}
\pdv{t}(\phi u) = D \grad \cdot (P \phi_\tau \grad u) + \phi \rho u (1-u)   \quad \text{in }  \Omega_B,
\label{eq:ddmpde}
\end{equation}
where $\phi_\tau=\phi+\tau$ and $\tau>0$ is small.
This PDE can be solved in $\Omega_B$, with a uniform discrete mesh set by the MRI image resolution (1mm in our case).
The boundary condition for $u$ in the extended domain $\Omega_B$ can be Dirichlet, Neumann, or periodic, as long as the computation boundary is far away from the brain boundary $\Omega$.
The phase field function $\phi$ can be obtained by evolving the Cahn-Hillard Equation for a short period of time:
\begin{equation}
  \begin{cases}
    \pdv{\phi}{t} = \grad \cdot \left( (\phi(1-\phi)) \cdot \grad (g'(\phi)-\epsilon^2 \Delta \phi) \right) \quad \text{on }\Omega_B\\
    \phi = 0 \quad \text{on } \partial \Omega_B\\
    \phi= 1_{\Omega} \quad \text{at } \{t=0\}
  \end{cases}
  \label{eq:cahn}
\end{equation}
where $g(\phi)=\phi^2(1-\phi)^2/4$ is a double-well potential, and $\epsilon$ is the width of the diffuse interface, chosen to be 3mm here. 
In our work, we solve \eqref{eq:cahn} and \eqref{eq:ddmpde} using a finite difference method (FDM) with standard central difference scheme and explicit Euler time stepping.

% We solve \eqref{eq:cahn} by finite difference method (FDM) on a regular grid. 
% We run the solver for 100 time steps to obtain $\phi$.

%%%%%%%%%%%%%%%%%%%%%%
\subsubsection{Scaling}\label{sss:scaling}

The MRI image shows the thresholded tumor cell density at an unknown time $\tend$.
% In single-time inference, the time between the onset of the tumor and the imaging is unknown.
It is impossible to infer $(D, \rho, \tend)$ simultaneously from a single time point:
doubling the time and halving $D$ and $\rho$ will give the same solution.
In \citet{konukogluImageGuidedPersonalization2010,ezhovLearnMorphInferNewWay2023,ezhovForloopAllYou2022}, 
time-independent parameters $\sqrt{D\tend}$ and $\sqrt{\rho \tend}$ are considered, and they suffice to predict the tumor cell density and infiltration pattern at the unknown time of imaging.
Additional information regarding the time scale would be needed to determine the precise values of $(D, \rho, \tend)$ 

In this work, we apply the idea from \citep{konukogluImageGuidedPersonalization2010,ezhovLearnMorphInferNewWay2023}, but with a different scaling technique.  
Let $\bD$ and $\brho$ be characteristic values of the diffusion coefficient and proliferation rate. Then, we can define the characteristic velocity $\bar{v} = \sqrt{\bD\brho}$. 
Given a length scale $\bL$, we can define a characteristic time $\bT = \bL/\bar{v}$. 

We can then define non-dimensional characteristic parameters $\cD$ and $\cR$:
\begin{equation}
  \cD = \frac{ \bD \bT }{\bL^2}, \quad
  \cR = \brho \bT.
\end{equation}
We also define the ratios of the diffusion coefficient and proliferation rate with respect to their characteristic values as
\begin{equation}
  \muD = \frac{D}{\bD} \frac{\tend}{\bT} , \quad
  \muR = \frac{\rho}{\brho}  \frac{t_{end}}{\bT} ,\quad
\end{equation}
By scaling the equation with $\bt= t/t_{end}$, $\bx = x/\bL$, we obtain the non-dimensional equation
\begin{equation}
  \begin{aligned}
    \pdv{u}{\bt} &= \muD \cD
  \grad \cdot ( P \phi \grad u) 
    +  \muR \cR \phi u (1-u), \\  
  \end{aligned}
  \label{eq:nondimpde}
\end{equation}
We also define the patient-specific characteristic solution $\bu$ as the solution of the PDE with $\muD = \muR = 1$. 

%
%% unused equation, for PPT
% \begin{equation*}
%   \begin{aligned}
%     \pdv{u}{\bt} &=  \underbrace{\frac{D_w}{\bD} \frac{\te}{\bT}}_{\muD}
%     \underbrace{ \frac{ \bD \bT }{\bL^2}}_{\cD} \grad \cdot ( P \phi \grad u) 
%     + \underbrace{\frac{\rho}{\brho}  \frac{t_{end}}{\bT} }_{\muR} \underbrace{\vphantom{\frac{\bD}{\bL} }\brho \bT}_{\cR} \phi u (1-u) \\  
%     % & =  \muD \cD \grad \cdot (\tilde{D}(\vb x) \grad u) + \muR \cR u(1-u)\\
%   \end{aligned}
% \end{equation*}
%

From our definitions, we have $\cD = {\sqrt{\ra}}/{\bL} = 1/\cR$.
Therefore $\bL$ and $\ra$ completely determine $\cD$, $\cR$, and hence $\bu$ accordingly. 
% As mentioned above, $\bL$ is patient-specific and $\ra$ will be constrained to generate a tumor of reasonable size regardless of the complex geometry of the brain: 
% 94\% of the tumor diameter at the time of diagnosis is less than 10cm\citep{palpanfloresAssessmentPreoperativeMeasurements2020,urbanskaGlioblastomaMultiformeOverview2014}.
% These considerations enable us to devise a simple procedure to estimate $\bL$ and $\ra$ by a grid search procedure using a surrogate model in a spherically symmetric geometry (see Sec~\ref{ss:workflow}).
We can use a grid search procedure to estimate $\bL$ and $\ra$ for each patient, which makes the characteristic parameters, and solution, patient-specific. The estimation uses a surrogate model in a spherically symmetric geometry (see Sec~\ref{ss:workflow}).
If our estimation of $\cD$ and $\cR$ is accurate, then we would expect $\muD$ and $\muR$ to be around 1.
Therefore instead of trying to learn parameters of different magnitudes, we instead learn parameters that are of order 1, which enhances the efficiency of the training algorithm.
In addition, the patient-specific characteristic solution $\bu$ is an excellent starting point for the PINN to learn patient specific parameters ($\muD$, $\muD$, etc ). 
Therefore, we use the FDM to obtain $\bufdm$, the FDM approximation of $\bu$, and use $\bufdm$ to pre-train the PINN. This will be discussed in Sec.~\ref{ss:workflow}. Because it is very efficient to obtain the patient-specific characteristic solution $\bufdm$ (e.g., 1-2 minutes of computation time), we tested whether $\bufdm$ itself provides a good approximation of the tumor cell density. We found that while the approximation using $\bufdm$ is surprisingly good, the fine-tuning step significantly improves accuracy (see \ref{ap:patient}, Table~\ref{t:charmetric} and Fig.~\ref{f:patchar}).

%n Table~\ref{t:charmetric}, we show the metrics for predictions using the characteristic parameters, that is, based on $\bufdm$. 
%For all cases, 
%$\muD = \muR = 1$, $m=1$, $A=0$, $\uct = 0.6$, $\ucf = 0.35$. 
%{The morphologies of these patient-specific characteristic tumors are shown in Fig.~\ref{f:patchar}}.

%%%%%%%%%%%%
\subsubsection{Multimodal Imaging Model}\label{sss:imaging}

% The segmentation $\vb y^s$ , $s\in {\rm T1Gd, FLAIR}$ assigns a label $y_i^s = 1$ to each voxel with visible tumor and $y_i^s = 0$ otherwise. 

We consider binary segmentations obtained from T1Gd, which highlights the tumor core, and FLAIR, which highlights the edema region in addition to the tumor core \citep{leMRIBasedBayesian2016,lipkovaPersonalizedRadiotherapyDesign2019,subramanianWhereDidTumor2020,tuncModelingGliomaGrowth2021,menzeGenerativeApproachImageBased2011,ezhovLearnMorphInferNewWay2023,konukogluExtrapolatingGliomaInvasion2010}.
Let $\segt$ be the indicator function of the tumor core and $\segf$ be the indicator function of the tumor core and edema region. 
We assume that the $\segt$ and $\segf$ are related to the tumor cell density $u$ through thresholding at $\uct$ and $\ucf$ respectively, that is, for $s\in \{\rm T1Gd, FLAIR\}$
\begin{equation}
    \segs(\vb x) = \mathbbm{1}_{\{u(\vb x)>\ucs\}}
\end{equation}

Following \cite{lipkovaPersonalizedRadiotherapyDesign2019}, we can also incorporate the 18-fluoride-fluoro-ethyl-tyrosine positron emission tomography (FET-PET) signal as additional data in the inference process. The FET-PET signal
measures the metabolic activity of the tumor cells and is  positively correlated to the tumor cellularity \citep{stockhammerCorrelationF18fluoroethyltyrosinUptake2008,hutterer18FFluoroethylltyrosinePET2013}.
However, FET-PET (hereafter referred to as FET) is not routinely available in the clinic.
Let $p$ be the normalized FET signal obtained by subtracting the patient-specific background activity, thresholding and normalizing to [0,1], and $\segfet$ be the indicator function of the region with positive FET signal. 
We assume that, in the region with positive normalized FET signal, there is a linear relationship between $p$ and $u$:
\begin{equation}
  p(\vb x) = (m u(\vb x) - A)\segfet(\vb x),
  \label{eq:fet}
\end{equation}
 where the intercept $A$ and the slope $m$ are patient-specific and unknown.
In \citet{lipkovaPersonalizedRadiotherapyDesign2019}, FET was found to be important to obtain accurate tumor cell density predictions.
Here, we show that FET is not necessary in our method.

%%%%%%%%%%%
\subsection{Workflow}\label{ss:workflow}
\begin{figure*}[!t]
  \centering
  \includegraphics[keepaspectratio,width=\textwidth]{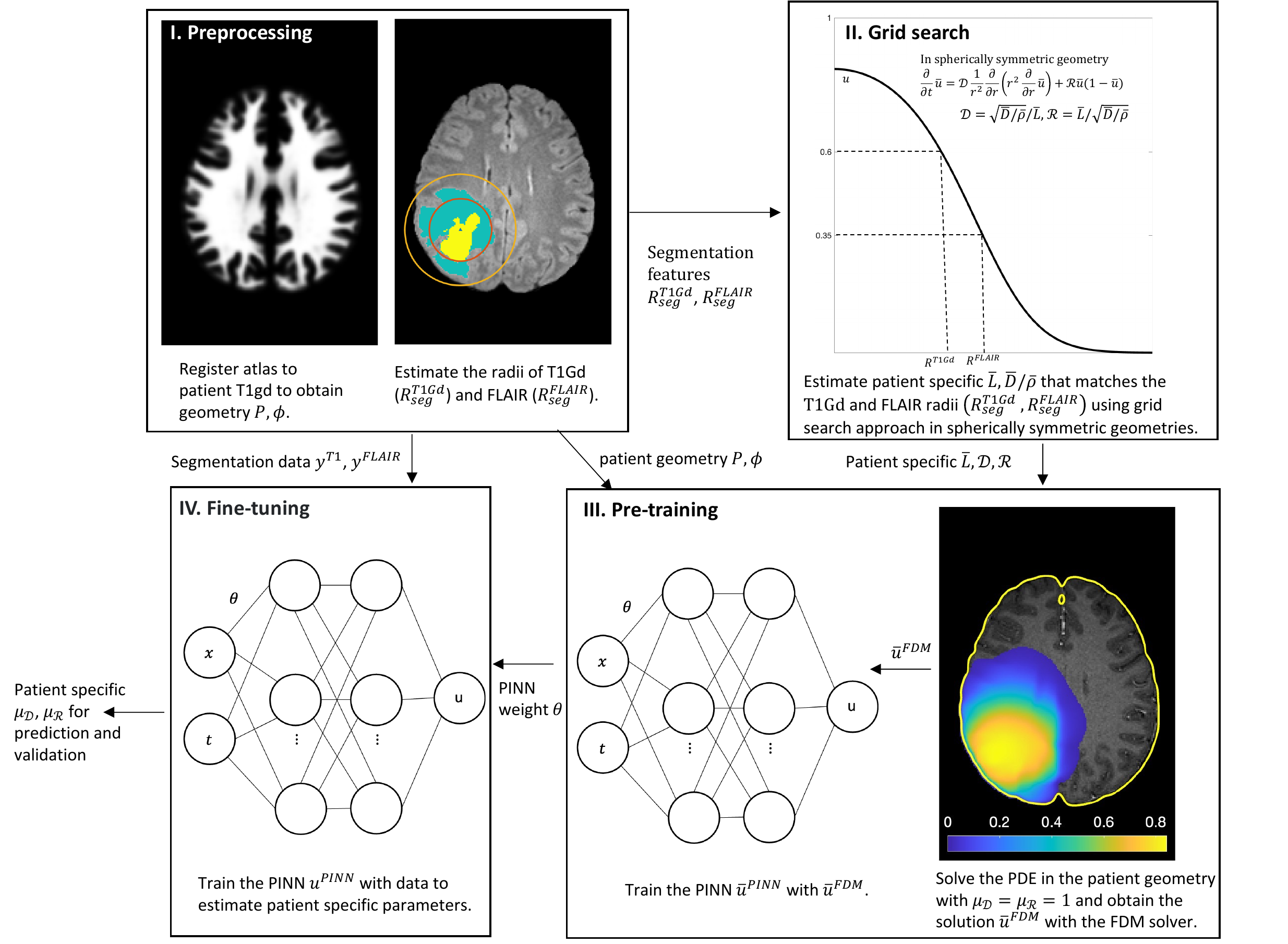}
  \caption{
    Workflow for patient-specific parameter estimation via PINN.
    \textbf{I. Preprocessing:} Register the brain atlas to the T1Gd patient scan. Solve the Cahn-Hilliard equation to acquire the geometry. Compute centers and radii of T1Gd and FLAIR segmentations ($\rt_{\rm seg}$ and $\rf_{\rm seg}$).
    \textbf{II. Grid Search:} In a spherically symmetric geometry, use a grid search algorithm to identify patient-specific characteristic values of $\ra$ and $L$ such that the radii derived from the PDE solution are close to  $\rt_{\rm seg}$ and $\rf_{\rm seg}$. 
    \textbf{III. Pre-training:} Solve the PDE using a Finite Difference Method (FDM) with the patient-specific characteristic parameters (and $\muD=\muR=1$) in the patient geometry to obtain a characteristic solution $\bufdm$. Train the PINN to solve the PDE using $\bufdm$ as data.
    \textbf{IV. Fine-Tuning:} Use the segmentation data to fine-tune the PINN and learn the patient specific parameters $\muD$ and $\muR$. These estimated parameters are used for tumor cell density predictions.
  }
  \label{f:workflow}
\end{figure*}

This section outlines the step-by-step workflow of our parameter estimation process, as visualized in Fig.~\ref{f:workflow}.
%% overall idea 
% The overall idea is that PINN is far away from the PDE solution at random initialization, which would lead to a long training time or might even failure \citep{wangWhenWhyPINNs2022}.
% However, based on the MRI data, we can quickly obtain an approximate parameter such that the solution of the PDE is of similar size as the segmentation.
% The solution can also be quickly obtained by FDM, as this only needs to be done once for each patient.
% Therefore, we can first train the PINN to approximate this characteristic solution, which would be ``closer'' to the optimal solution, and then fine-tune the PINN with the segmentation data to obtain the patient-specific parameters.
In the preprocessing step, we first register the atlas to the patient T1Gd and obtain the geometry $P(\vb x)$, and solve Eq.~\eqref{eq:cahn} to obtain the phase field function $\phi$. 
Next from the segmentation, we compute the centroid of the tumor core, and the maximum radii of $\segf$ and $\segt$, denoted as $\rf_{\rm seg}$ and $\rt_{\rm seg}$, which serve as features from the images. 

Next, we estimate the characteristic parameters $\ra$ and $\bL$ by a grid search procedure.
If we ignore the complex geometry of the brain, i.e. $P(\vb x)=1$, then in spherical coordinates the characteristic PDE reduces to a one-dimensional equation $\busph(r,t)$ with Neumann boundary conditions, which can be solved easily:
\begin{equation}
  \pdv{\busph}{t} = \cD \frac{1}{r^2} \pdv{r} \left( r^2 \pdv{r} \busph \right)  +  \cR \busph (1-\busph)
  \label{eq:pde1d}
\end{equation}
For samples in $\ra \in [0.1, 1]$ and $\bL \in [10,90]$ (which generate spherical tumors with radii smaller than 120mm), we solve the simplified PDE to find the radii $\rf_{sph}$ and $\rt_{sph}$ at which the tumor cell density is equal to characteristic values of the thresholds, which are taken to be $\bucf=0.35$ and $\buct=0.6$, respectively, motivated by results in \cite{lipkovaPersonalizedRadiotherapyDesign2019}.
The pair of $\ra$ and $L$ that results in $\rf_{sph}$ and $\rt_{sph}$ that are closest to $\rf_{\rm seg}$ and $\rt_{\rm seg}$ will be used as the characteristic parameters.
The details of this procedure are given in \ref{ap:char}.
This procedure is very fast as the PDE is one-dimensional, independent of the patient data, and the quantities $\rf_{sph}$ and $\rt_{sph}$ can be precomputed and stored as a look-up table.
In addition, we do not need to perform a very refined grid search, because the actual parameters will be fine-tuned by the PINN with patient data.

With the patient specific $\cD$ and $\cR$, we can solve the original PDE \eqref{eq:nondimpde} with $\muD=\muR=1$ using the FDM in the patient's geometry to obtain the characteristic solution $\bufdm$. 
We then train the PINN to approximate $\bufdm$.
We call this step pre-training as it prepares the PINN for estimating the patient-specific parameters by initializing the weights of the PINN so that the PINN approximates a PDE solution instead of being a random function without PDE structure. 
In our experience, this makes the training process much more efficient and robust.

In the final stage, we use the segmentation data to fine-tune the weights of the PINN and learn the patient specific parameters $\muD$ and $\muR$. 
These estimated parameters can be used for tumor predicting the tumor cell densities.

%In the next Section, we define the loss function for pre-training and fine-tuning of the PINN.

%%%%%%%%%%%
\subsection{Loss functions} \label{ss:loss}

In this section, we define the loss functions for pre-training and fine-tuning of the PINN.
For notational simplicity, let $u(\vbx,t)$ denote the function represented by the PINN, with $x$ and $t$ already normalized.
To enforce the initial condition, we set $u(\vbx,t) = t u^{NN} (\vbx,t) + u_0(\vbx)$, where $u^{NN}$ is a fully connected neural network. 
Define the differential operator $\mathcal{F}$ of the PDE \eqref{eq:nondimpde} as
\begin{equation}
  \mathcal{F}[u] = \muD \cD \left(\grad ( P \phi) \cdot \grad u +  P \phi \laplacian u\right) + 
  \muR \cR \phi u (1-u) - \pdv{u}{\bt}
\end{equation}
Here we write the PDE in non-divergence form, since the geometry $P \phi$ and its gradient $\nabla (P \phi)$, which is approximated by finite differences using data from the pixels in the MRI scans, are known data. 

Let $\{\xres_i,t_i\}_{i=1}^{\Nres}$ be the collocation points for the residual loss function.
The residual loss function is defined as
\begin{equation}
  \lres = \frac{1}{\Nres}\sum_{i=1}^{\Nres} \left(\mathcal{F}[u](\xres_i, t_i)\right)^2,
\end{equation}
Because the solution is changing rapidly at early times and the tumor grows from the center, 
we sample the collocation points for PDE loss densely at early times and in the center of the tumor. 
This was found to work better than other spatial distributions of the collocation points, including adaptive sampling \citep{luDeepXDEDeepLearning2021} 
(see \ref{ap:xdist} for details).

To facilitate training of the characteristic PINN, we also minimize the Mean Squared Error (MSE) with respect to the characteristic solution $\bufdm$
\begin{equation}
  \lchar = \frac{1}{\Nres}\sum_{i=1}^{\Nres} \left(\phi(\xres_i) u(\xres_i, t_i) - \phi(\xres_i)\bufdm(\xres_i, t_i) \right)^2.
  \label{eq:char}
\end{equation}
Thus the total loss function for pre-training is
\begin{equation}
  \ltot = \lres + \lchar.
\end{equation}

For fine-tuning the PINN, we replace $\lchar$ by loss functions based on the imaging data.
Let $\{\xdat_i\}_{i=1}^{\Ndat}$ be the spatial collocation points for the data loss.
Note that all the imaging data are provided at the normalized time $t=1$. The points
$\xdat$ are sampled uniformly at pixel resolution.
We define the predicted segmentation obtained by thresholding the cell density at $u_c^s$:
\begin{equation}
  H_{u_c^s}(x) = \left[1+\exp{-a (\phi(\vbx)u(\vbx,1) -u_c^s)}\right)^{-1}
\end{equation}
which is a smoothed heaviside function in [0,1] (we set $a=20$). 
The segmentation loss function is defined as
\begin{equation}
  \begin{split}
    \lseg = \frac{1}{\Ndat} \sum_{i=1}^{\Ndat} \left( H_{\ucf}(\xdat_i) - \segf(\xdat_i) \right)^2 + \\ \left( H_{\uct}(\xdat_i) - \segt(\xdat_i) \right)^2
  \end{split}
\end{equation}
where $\ucf$ and $\uct$ are trainable parameters.
If the FET data is used, the loss function for FET signal is defined as
\begin{equation}
  \lfet = \frac{1}{\Ndat} \sum_{i=1}^{\Ndat} \left( m \phi(\xdat_i)u (\xdat_i,1) - A - p(\xdat_i) \right)^2 \segfet(\xdat_i)
\end{equation}
where $m$ and $A$ are trainable parameters.
% In addition, the loss on the boundary condition
% \begin{equation}
%   \lbc = \frac{1}{\Nbc} \sum_{i=1}^{\Nbc} \left( u(\xbc_i, t_i)  \right)^2
% \end{equation}

In the fine tuning stage, we also restrict the range of the trainable parameters:
$\muD\in[0.75, 1.25]$,
$\muR\in[0.75, 1.25]$,
$\ucf\in[0.2,0.5]$,
$\uct\in[0.5,0.8]$,
$m\in[0.8,1.2]$,
$A\in[0,1]$.
As discussed earlier, we expect $\muD$ and $\muR$ to be close to 1.
The ranges for other parameters are based on the priors in \citep{lipkovaPersonalizedRadiotherapyDesign2019}.
The constraints are enforced by adding the quadratic penalty losses: if a parameter $\beta\in[\beta_{\min}, \beta_{\max}]$, then 
\begin{equation}
  \mathcal{L}_{\rm \beta} = \min(0, \beta-\beta_{\min})^2 + \max(0, \beta-\beta_{\max})^2
  \label{eq:reg}
\end{equation}

In all, the total loss for fine-tuning with segmentation and FET data (if used) is
\begin{equation}
  \ltot = \lres + \wseg \lseg + \wfet \lfet + \sum_{\beta \in \Theta} \mathcal{L}_{\rm \beta},
  \label{eq:totalloss}
\end{equation}
where $\Theta = \{\muD, \muR, \icx, \ucf, \uct, m, A\}$ are trainable parameters, and $\wseg$ and $\wfet$ are the weights of the segmentation and FET loss respectively. 
If FET data is not used, then $\wfet=0$.
In this work we set $\wseg=1e-3$ and $\wfet=1e-2$ if FET data is used, while all the other losses have unit weights.
These weights are chosen such that the losses are of similar magnitude,
such that the training is not dominated by one particular loss. 
Although we tried other choices of the weights, including adaptively changing them in time {\citep{groenendijkMultiLossWeightingCoefficient2020,madduInverseDirichletWeighting2022,chenGradNormGradientNormalization2018,wangUnderstandingMitigatingGradient2021}}, we found empirically that this choice of weights works well for all our numerical experiments.
See \ref{ap:training} for details on training the neural network and for examples of the dynamics of loss functions during training. See \ref{ap:datatype} for different choices of the data loss.

%%%%%%%%%%%%%%%%%%%%%%
\section{Results}\label{s:results}

% \subsection{Overview}
We start by presenting an overview of the results and our findings.
In this work we use 3 different datasets, including data from 24 patients, to test and validate our approach:
\begin{itemize}
  \item Synthetically-generated data, denoted as S1-S8:
  We specify the ground truth (GT) parameters, solve the PDE by the FDM to obtain the cell density, 
  add spatially correlated noise to the density, and generate synthetic segmentations and FET distributions.
  \item Data from 8 patients (P1-P8), primary tumors and first detected recurrence, from \citep{lipkovaPersonalizedRadiotherapyDesign2019}. All the patients have T1Gd and FLAIR scans at recurrence.
  \item  Data from 16 patients (Q1-Q16), primary and first detected recurrence, from the Klinikum Rechts Der Isar repository. At recurrence, patients Q1-Q3 have both T1Gd and FLAIR scans, the others only have T1Gd scans.
  \end{itemize}

We test parameter estimation using segmentions of T1Gd and FLAIR scans and FET-PET (denoted as \petseg{}), and using segmentations alone (denoted as \segonly{}).
The difference between the two lies in the loss functions used in the fine-tuning stage: \petseg{} includes the FET loss $\lfet$, in addition to the segmentation loss $\lseg$.

We demonstrate that using segmentation alone achieves results comparable to those using segmentation and FET, in contrast to the results in \citep{lipkovaPersonalizedRadiotherapyDesign2019} where FET was needed to more accurately determine spatial  distributions of the cells, particularly near the tumor center. This finding is noteworthy as FET-PET data is not widely available from the clinic, thereby accentuating the practicality of our approach.

Furthermore, we examine two methods to predict cell density after fine-tuning: $\ufdm$, derived from solving the PDE with estimated parameters using the FDM, and $\upinn$, obtained by evaluating the PINN. 
We show that $\ufdm$ is more accurate in terms of solving the PDE as $\upinn$ is influenced by the noisy data via the loss function.
Additionally, the FDM can be used to solve the PDE to any non-dimensional time.
However, extending the PDE solution to future times with the PINN requires re-sampling and extending the collocation points, and retraining the network with the PDE loss, which tends to be more time consuming.

%%%%%%%%%%%%%%%%%%%%%%
\subsection{Synthetic data} \label{ss:synthetic}
\begin{figure}[!t]
  \centering
  \includegraphics[keepaspectratio,width=\columnwidth]{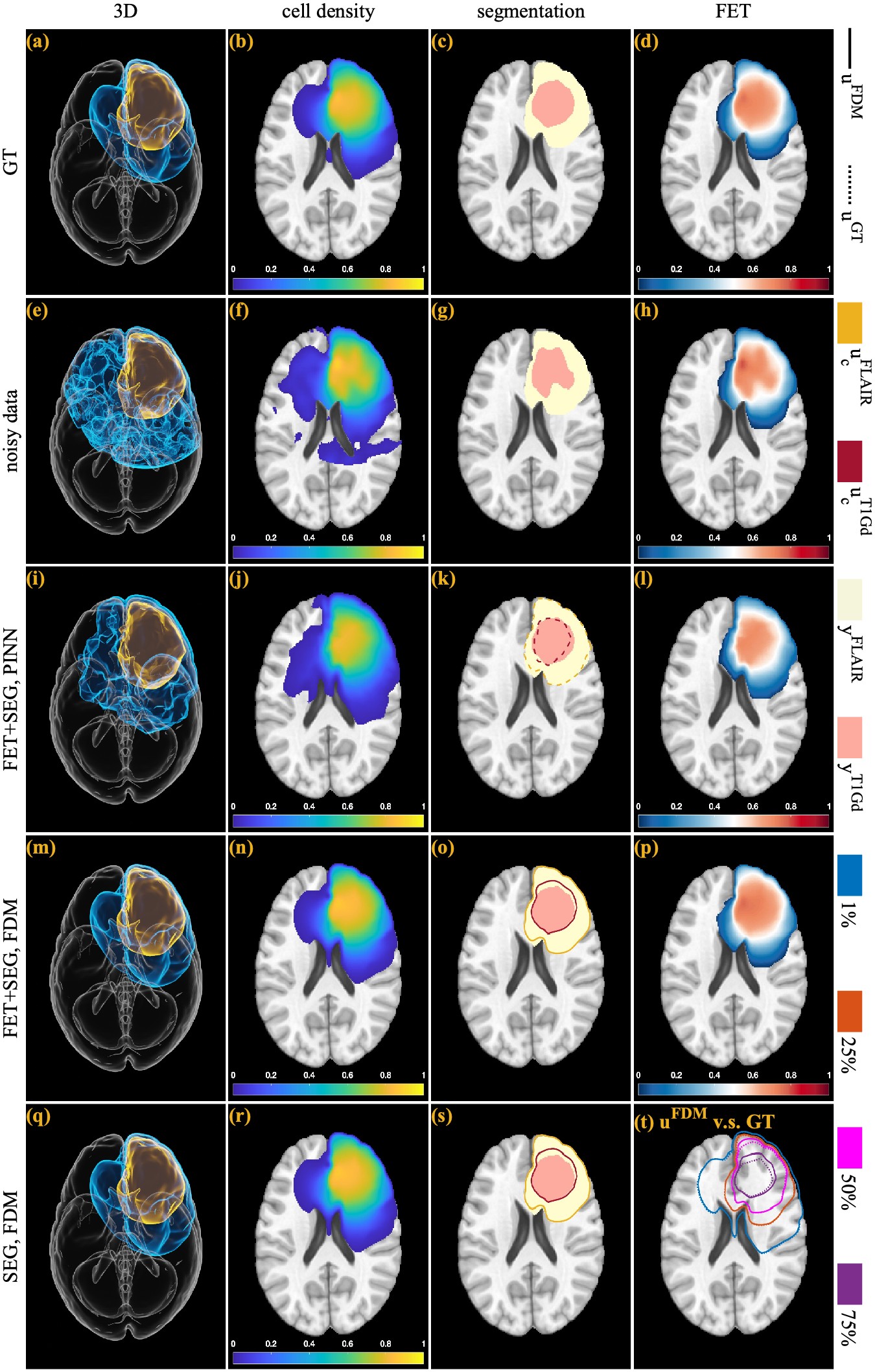}
  \caption{
    Validation using synthetic data.
 Row 1:  (a)-(d). ground truth (GT) data; Row 2: (e)-(h). Training data-- GT with correlated noise;
 Row 3: (i)-(l). Parameter estimation using 
  noisy, synthetic  \petseg{}  data and predictions by evaluating the PINN solution, $\upinn$; Row 4: (m)-(p). Parameter estimation using 
  noisy \petseg{} data and predictions using the FDM solution of the PDE, $\ufdm$,
  with inferred parameters; Row 5: (q)-(t). Parameter estimation using only the
  noisy \segonly{} data and predictions using $\ufdm$.
 (t): Comparison of contours at 1\%, 25\%, 50\%, and 75\% between $\ufdm$ (solid) and GT $u$ (dotted).
    Column 1: 3D isosurfaces of tumor cell density at 1\%  and $\segf$;
    Column 2: Cell densities;
    Column 3: Synthetic segmentations mimicking T1Gd and FLAIR data;
    Column 4: Synthetic FET-PET distributions.
 Further, (k), (o) and (s) show comparisons of predicted segmentations (lines--dashed, solid) with GT segmentations (filled).
%    
%    
%    Row 1: Ground truth.
%    Row 2: Training data with spatially correlated noise (see Appendix Fig.~\ref{f:noise}).
%    Row 3: \petseg{} outcomes using $\upinn$.
%    Row 4: \petseg{} outcomes using $\ufdm$.
%    Row 5: \segonly{} outcomes using $\ufdm$. 
%    
%    (t): Comparison of contours at 1\%, 25\%, 50\%, and 75\% for $\ufdm$ and GT $u$.
 Comparing rows 3 and 4, $\upinn$ is less accurate than $\ufdm$  as a solution of the PDE with the inferred parameters
 because $\upinn$ is more strongly influenced by the noisy data.
 Notably, the estimated cell densities from the  $\ufdm$ using \petseg{} and  \segonly{} in rows
 4 and 5, respectively, are very similar and yield accurate approximations of the GT data.
    }
  \label{f:syndetail}
\end{figure}

Figure~\ref{f:syndetail} presents detailed results for a synthetic tumor (labeled S5); the ground truth and inferred parameters are shown in Table~\ref{t:syn}. 
The results for the other synthetic tumors are shown in the Supplementary Material.
The first row shows the Ground Truth (GT) data: 3D isosurfaces of tumor cell density at 1\% and 30\%, cell density, synthetic segmentation, and synthetic FET signal. 
The synthetic cell density, denoted as  $\ugt$, is obtained by solving the PDE with the GT parameters ($\muD,\muR,\icx$) using the FDM (the results would be similar if we solve the PDE using the PINN).
The synthetic segmentation is obtained by thresholding $\ugt$ at $\uct$ and $\ucf$.
The synthetic FET signal is formulated as $p = (m\ugt-A)\segfet$, and $\segfet$ is the region where $m\ugt-A>0$.
The second row presents the training data generated by introducing spatially correlated noise to the GT, 
(see \ref{ap:synthetic} and Figure~\ref{f:noise} for further details of the noise).
The segmentations from the noise-perturbed solutions deviate noticeably from the GT.
In the third and fourth rows, we show the results of \petseg{} using $\upinn$ in row 3 and $\ufdm$, with the estimated parameters, in row 4.
% We see that $\ufdm$ fits the ground truth data better than $\upinn$.
Although the inferred segmentations for the PINN (dashed, Fig.~\ref{f:syndetail}(k)) are closer to the ground truth than for the FDM (solid, Fig.~\ref{f:syndetail}(o)), the cell density predicted by $\ufdm$ overall is a better fit to the ground truth data than is
$\upinn$  (the MSEs for $\ufdm$ and  $\upinn$ are $3.8\times 10^{-5}$ and $1.4\times 10^{-4}$, respectively).
This is because the PINN is more influenced by the noisy data via the data loss term in the loss function.
Therefore $\ufdm$ will be used for our predictions.
Comparing the fourth and fifth rows, the estimated cell density using \petseg{} is very similar to that using \segonly{} alone, with both being close to the GT because the estimated parameters are accurate (\ref{ap:synthetic} Table~\ref{t:syn}).
As \segonly{} simulations do not provide estimated FET signals, Figure~\ref{f:syndetail}(t) directly compares the contours of $\ufdm$ using \segonly{} with GT contours at 1\%, 25\%, 50\%, and 75\%, and shows there is very good agreement with the ground truth data.

\begin{figure}[!t]
  \centering
  \includegraphics[keepaspectratio,width=\columnwidth]{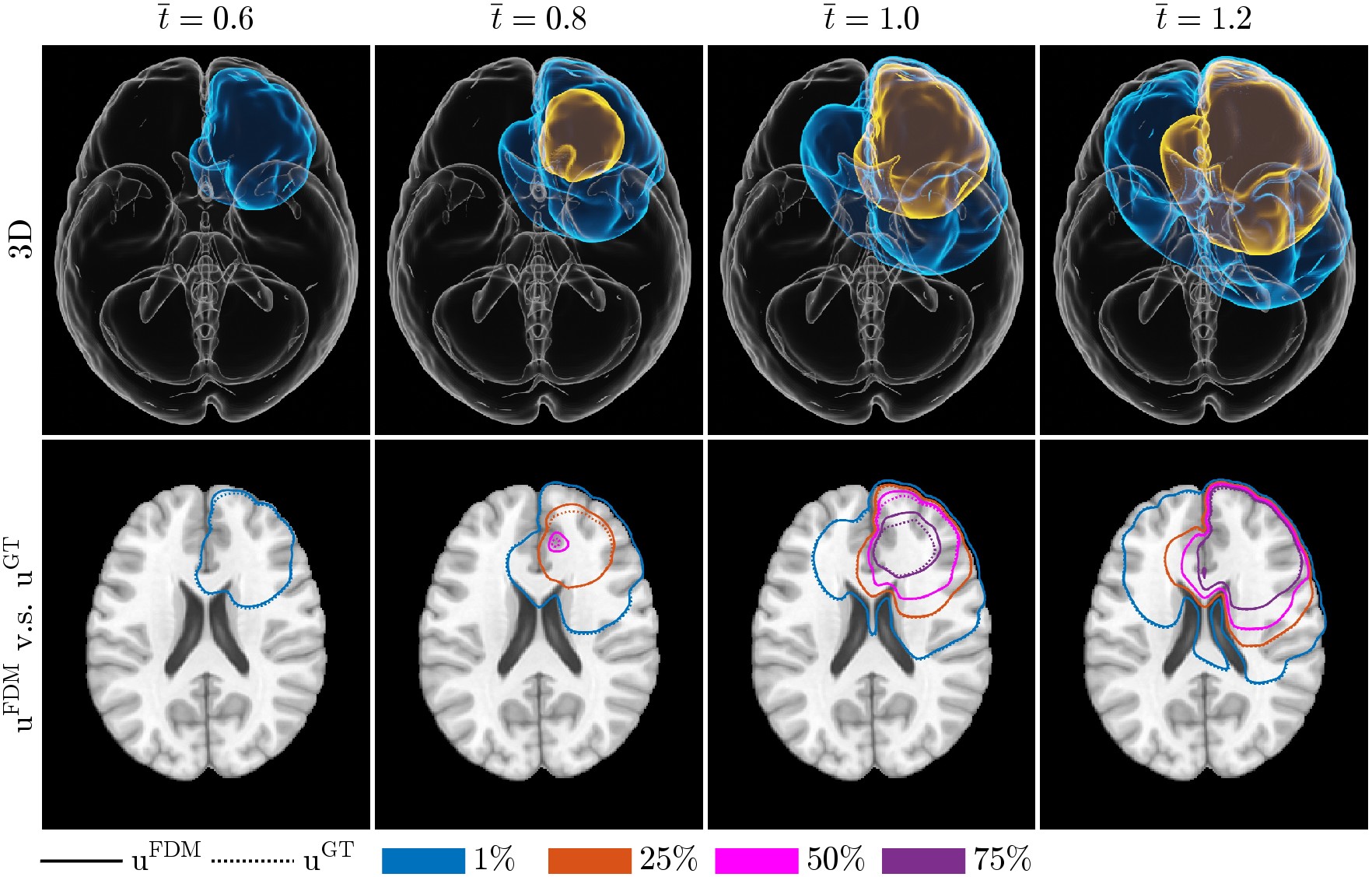}
  \caption{ Predicted tumor morphologies at times different from that of training data.
  Morphologies are shown at the normalized times $\bar{t}$ = 0.6, 0.8, 1.0, 1.2 and are
   obtained using $\ufdm$ with parameters inferred from the synthetic data in Fig. \ref{f:syndetail}. 
  Only \segonly{} data is used for inference and the training data corresponds to the normalized time $\bar{t}=1$.
  Row 1: 3D isosurfaces of $\ufdm$ at 1\% and 30\%
  Row 2: Comparision of predicted contours (solid; $1\%,~25\%,~50\%$ and $75\%$) with GT  (dotted).
  }
  \label{f:syndynamics}
\end{figure}

% As shown in Fig.~\ref{f:syndetail}, it is better to use $\ufdm$ to predict tumor density than $\upinn$. 
The PDEs with the inferred parameters can be used to predict tumor evolution at arbitrary times, before or after the time of imaging.
Directly evaluating the PINN at time beyond the range of the temporal domain used in training tends to yield poor results,
although this can be remedied by re-sampling collocation points and re-training the network with the PDE loss, or by novel methods to enhance the extrapolation capabilities of PINN \citep{kimDPMNovelTraining2020}.
It should be noted that our method only permits the use of normalized time as the time scale remains unknown.
Figure~\ref{f:syndynamics} demonstrates the predicted dynamics of tumor cell density $\ufdm$ at normalized times $\bar{t}$ = 0.6, 0.8, 1.0, 1.2, with data provided at $\bar{t}=1$.
The first row shows the 3D isosurfaces of $\ufdm$ at 1\% and 30\%.
The second row shows the predicted contours and those of the GT.
The predicted dynamics of tumor growth align closely with the GT because the estimated parameters are close to the GT parameters.

%%%%%%%%%%%
\subsection{Patient tumors} \label{ss:patient}
\begin{figure}[!t]
  \centering
  \includegraphics[keepaspectratio,width=\columnwidth]{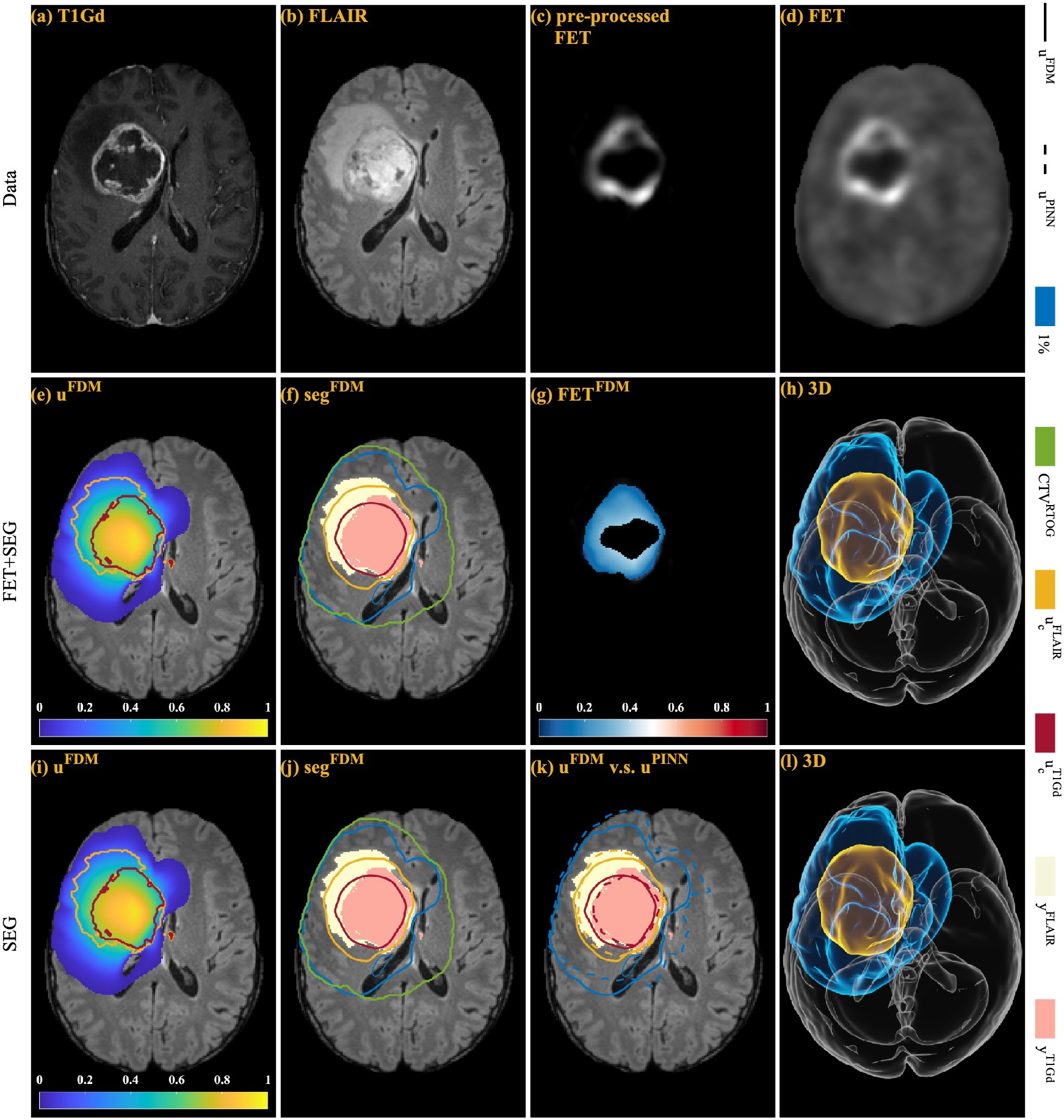}
  \caption{
Predicted patient-specific tumor morphologies, cell densities and FET-PET  distributions from Patient 5 (see text).
Row 1: Medical images:
(a) T1Gd;
(b) FLAIR;
(c) Post-processed FET-PET (see text);
(d) Original FET-PET data.
Row 2:  $\ufdm$ predictions using \petseg{} data. Row 3:  $\ufdm$ predictions using \segonly{} data only.
(e) and (i): Predicted cell density distributions from $\ufdm$  together with boundaries of the segmentations (curves)
from the MRI images. (f) and (j): Comparisons of the contours of $\ufdm$ at the 
inferred segmentation thresholds (curves) with MRI segmentations (filled). Margins for $\ctvrtog$ (green) and $\ctvp$ (blue);
(g) predicted FET-PET  distribution (see text).
(h) and (l): 3D visualization of 1\% and $\ucf$ isosurfaces.
(k) Comparison of  1\%, $\ucf$, and $\uct$ contours of $\ufdm$ (solid) and $\upinn$ (dashed).
}
\label{f:patdetail}
\end{figure}

In this section, our method is applied to data from 24 patients: P1-P8 and Q1-Q16. As an example, Figure~\ref{f:patdetail} demonstrates the results of the inference for patient P5 using \segonly{} and \petseg{} as data.
The estimated parameters are given in Table~\ref{t:patparammetric}.
The first row presents the preoperative scans (a) T1Gd (b) FLAIR, (c) the pre-processed FET (as described in \ref{sss:imaging}) and (d) the original FET data.
The second and third rows exhibit the results for $\ufdm$ using \petseg{} and \segonly{} data, respectively.
Upon comparing columns 1, 2, and 4, it is evident that the estimations using \petseg{} and \segonly{} result in similar predicted tumor cell densities, and the predicted segmentations align closely with the actual segmentations. 
The Radiation Therapy Oncology Group (RTOG) Clinical Target Volume (CTV), denoted as $\ctvrtog$, is defined as the 2 cm margin around the FLAIR segmentation (\citet{gilbertDoseDenseTemozolomideNewly2013}).
The personalized CTV, denoted as $\ctvp$, is defined by thresholding the predicted tumor cell density at $\ufdm = 1\%$ across all patients (blue contours in Figs.~\ref{f:patdetail} (f) and (j)).
This threshold was selected so that the efficiency of the $\ctvp$ is comparable to that of the $\ctvrtog$ (green contours in Figs.~\ref{f:patdetail} (f) and (j)). 
Figures~\ref{f:patdetail}(f) and (j) reveal that both $\ctvp$ and $\ctvrtog$ encompass the visible tumor, although $\ctvp$ is smaller in size compared to $\ctvrtog$.
In Figure~\ref{f:patdetail}(g), the predicted FET signal using the \petseg{} data is displayed, and is highly correlated with the actual FET signal (see \ref{ap:patient} Table~\ref{t:patparammetric}).

%While rigid registration works well in many cases, some errors may arise 
% near the ventricles, especially when there is significant deformation of the brain due to the tumor.
%It remains an active area of research to design registration methods that robustly account for significant distortions due to tumor mass effect and post-surgical resection \cite{jana}.
%
%Also we just use one atlas, which might not be a good atlas. In Biro's paper (Ensemble Inversion for Brain Tumor Growth Models With Mass Effect), they have a collection of templates/atlas, and they pick a few templates that have similar ventricle size as the patient to solve the inverse problem.

Figure~\ref{f:patdetail}(k) compares the contours of $\ufdm$ (solid) and $\upinn$ (dashed) using \segonly{} data.
A comparison of density distributions is shown in \ref{ap:patient} Fig.~\ref{f:patpinnvsfdm}.
As observed in the synthetic data, the predicted segmentation from $\upinn$ aligns more closely with the actual segmentation than $\ufdm$ does, while the 1\% contour is somewhat noisier than that from $\ufdm$.
However, a close alignment with the actual segmentation is not necessarily advantageous as it suggests that $\upinn$ is fitting the data rather than effectively solving the PDE.
This behavior is inherent in the PINN as the PDE loss merely imposes a soft constraint on the PINN.
In particular, when the model cannot fully account for the data, PINN might sacrifice the PDE loss to fit the data, as long as the total loss is reduced.
Our synthetic data experiments indicate that even though $\upinn$ at time 1 is influenced by the noise, the estimated parameters remain reliable.
Therefore, despite the absence of a ground truth with patient data, we advocate for the use of $\ufdm$ for predictions, as it solves the PDE without being swayed by the noise and the PDE errors, and provides less noisy segmentations and 1\% contours, which are important for predicting infiltration.

\begin{figure}[!t]
  \centering
  \includegraphics[keepaspectratio,width=\columnwidth]{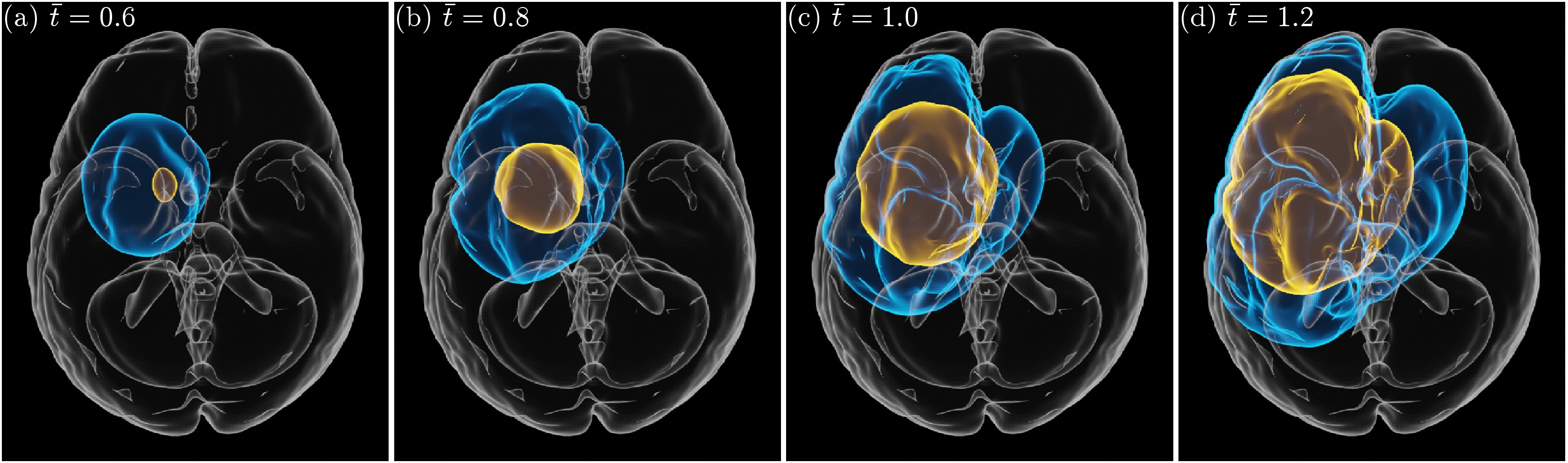}
  \caption{ Predicted dynamics of tumor from Patient 5 using the FDM at normalized times $\bar{t}$ = 0.6, 0.8, 1.0, 1.2.. 
  The data is provided at $\bar{t}=1$.  The 3D isosurfaces of $\ufdm$ at 1\% and 30\% are shown.
  }
  \label{f:patdynamics}
\end{figure}

As in the synthetic data case, the evolution of patient tumors in normalized time can be studied.
Fig.~\ref{f:patdynamics} shows a 3D visualization of the predicted dynamics of the tumor growth $\ufdm$ from patient P5 at normalized times $\bar{t}$ = 0.6, 0.8, 1.0, 1.2 (recall that the data is provided at $\bar{t}=1$).

\begin{figure*}[!t]
  \centering
  \includegraphics[keepaspectratio,width=\textwidth]{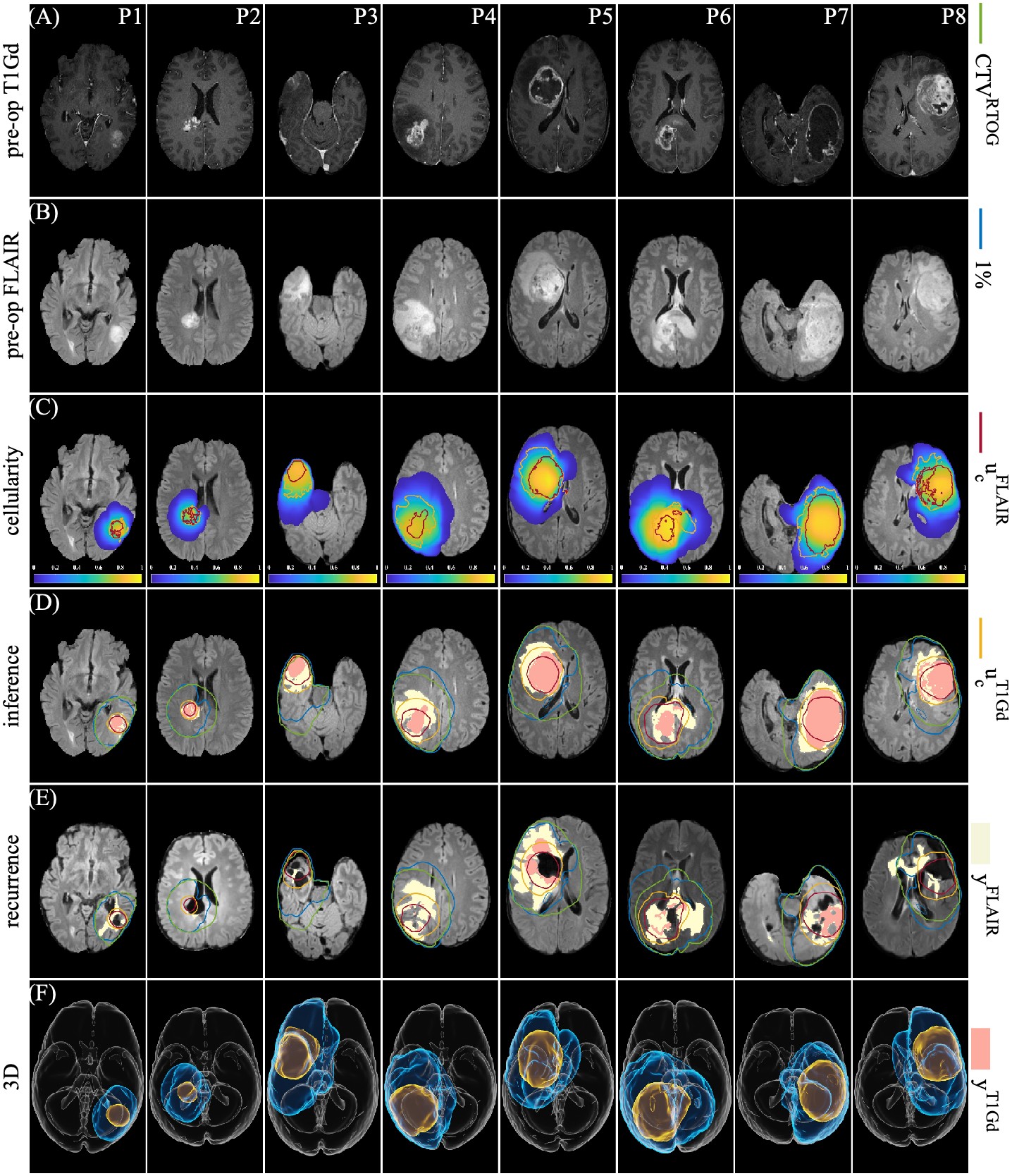}
  \caption{
Parameter Estimation Results for Patients P1-P8 using \segonly{}.
(a) Pre-operative T1Gd.
(b) Pre-operative FLAIR.
(c) Superposition of predicted tumor cell density $\ufdm$ on actual segmentations: T1Gd (yellow line) and FLAIR (red line).
(d) Overlap of inferred segmentations: $\ucf$ (yellow line) and $\uct$ (red line) on actual segmentations: T1Gd (beige fill) and FLAIR (pink fill); Margins for $\ctvrtog$ (green) and $\ctvp$ (blue).
(e) Tumor recurrence, compared with with margins for $\ctvrtog$ (green) and $\ctvp$ (blue).
(f) 3D reconstructions of $\ctvp$ and $\ucf$ isosurfaces.
On average, pre-training took 37.9 (std 5.0) minutes, while fine-tuning was completed in 4.5 (std 0.1) minutes.
}
\label{f:patseg}
\end{figure*}

%% about the figure
Figure~\ref{f:patseg} shows the results for patients P1-P8 using \segonly{} data.
% The segmentations based on Fig.~\ref{f:patseg} (a) T1Gd and (b) FLAIR are used for parameter estimation.
The estimated model parameters can be found in \ref{ap:patient} Table~\ref{t:patparammetric}.
The T1Gd and FLAIR images are shown in Figs.~\ref{f:patseg} (A) and (B),
and the predictions are shown in Figs.~\ref{f:patseg} (C-F).
%% about recurrence
All patients received standard treatment, which consists of immediate tumor resection, followed by combined radiotherapy and chemotherapy, and are monitored for recurrence regularly.
Fig.~\ref{f:patseg} (e) shows the post-operative FLAIR scan with segmentations of the first detected tumor recurrences.
The postoperative scans are registered to the preoperative T1Gd scans by rigid transformations \citep{lipkovaPersonalizedRadiotherapyDesign2019}.
% as the resection might significantly change the brain anatomy and results in large registration error in nonrigid registration.
The figures also plot the $\ctvp$ (blue), indicative of the predicted tumor infiltration pathways, and $\ctvrtog$ (green) for comparison.
% For relative small recurrence tumor, as in P1-P3, both $\ctvp$ and $\ctvrtog$ cover the recurrence tumor.
All the $\ctvp$ and $\ctvrtog$ cover the recurrent tumor reasonably well.
%In  patients P2, P3, P5, $\ctvp$ spares more healthy tissue.
Further, for patient P5, $\ctvp$ follows the contour of the recurrent tumor.
% In P4, though $\ctvp$ is larger than $\ctvrtog$ in the top, we can see the recurrence tumor is also growing more aggressively towards the top.
For patient P8, we see infiltration into the other brain hemisphere beyond the $\ctvp$ and $\ctvrtog$. However, our predicted $\ctvp$ seems to capture the infiltration pathway.
The estimated parameters and the metrics, including DICE scores of the segmentations and correlations with FET signals, corresponding to Fig.~\ref{f:patseg} \segonly{} are reported in \ref{ap:patient}, Table~\ref{t:patparammetric}.
The results of \petseg{} are reported in \ref{ap:patient}, Fig.~\ref{f:patsegpet} and Table~\ref{t:patparammetric}.

\begin{figure*}[!h]
  \centering
  \includegraphics[keepaspectratio,width=\textwidth]{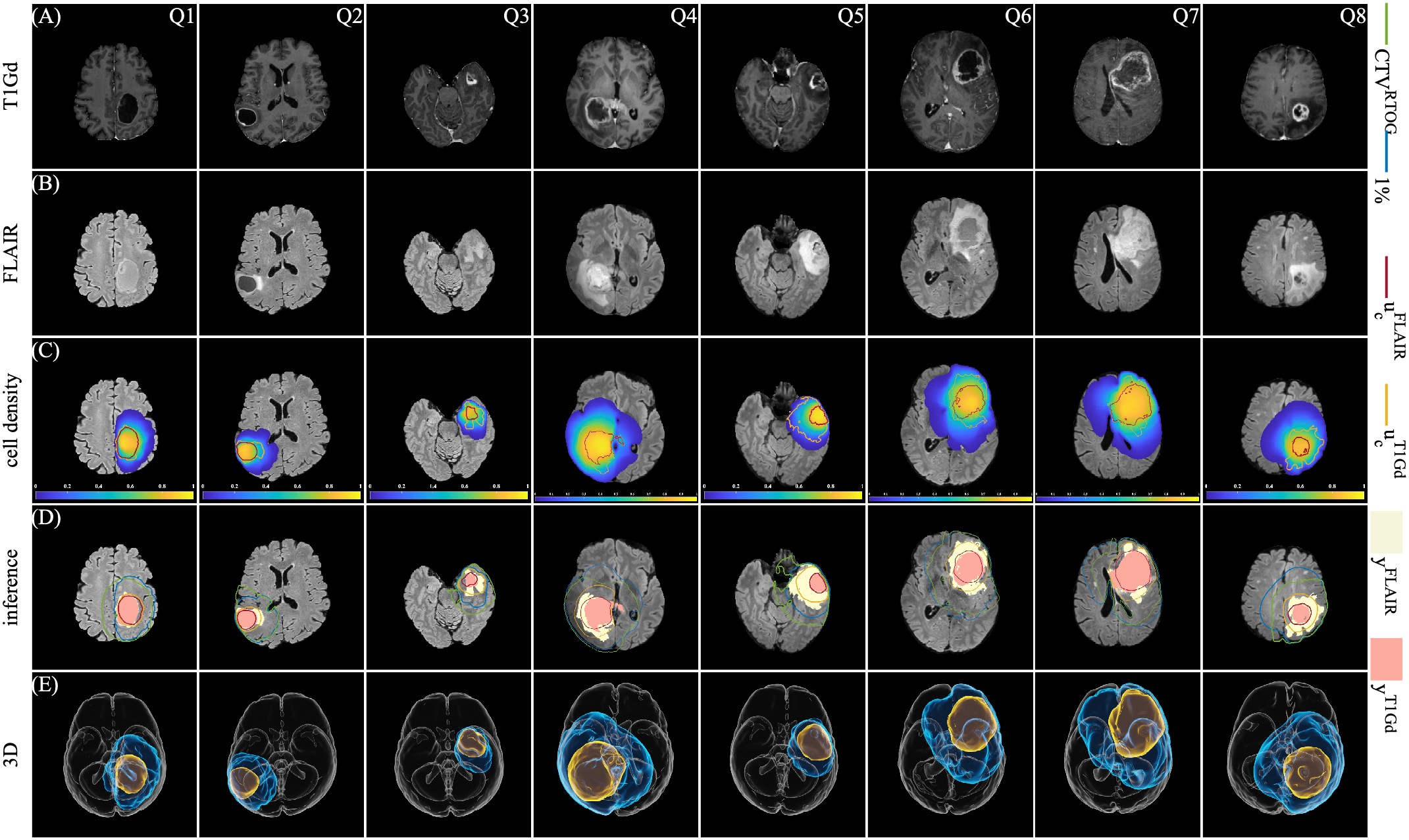}
  \caption{Parameter Estimation Results for Patients Q1-Q8 using \segonly{}. Recurrence results for Q1-Q3 are shown in \ref{ap:patient} Fig~.\ref{f:benerec}. Legends are the same as \ref{f:patseg}. Results using \petseg{} are shown in \ref{ap:patient} Fig.~\ref{f:addpetseg1}.}
  \label{f:addseg1}
\end{figure*}

\begin{figure*}[!h]
  \centering
  \includegraphics[keepaspectratio,width=\textwidth]{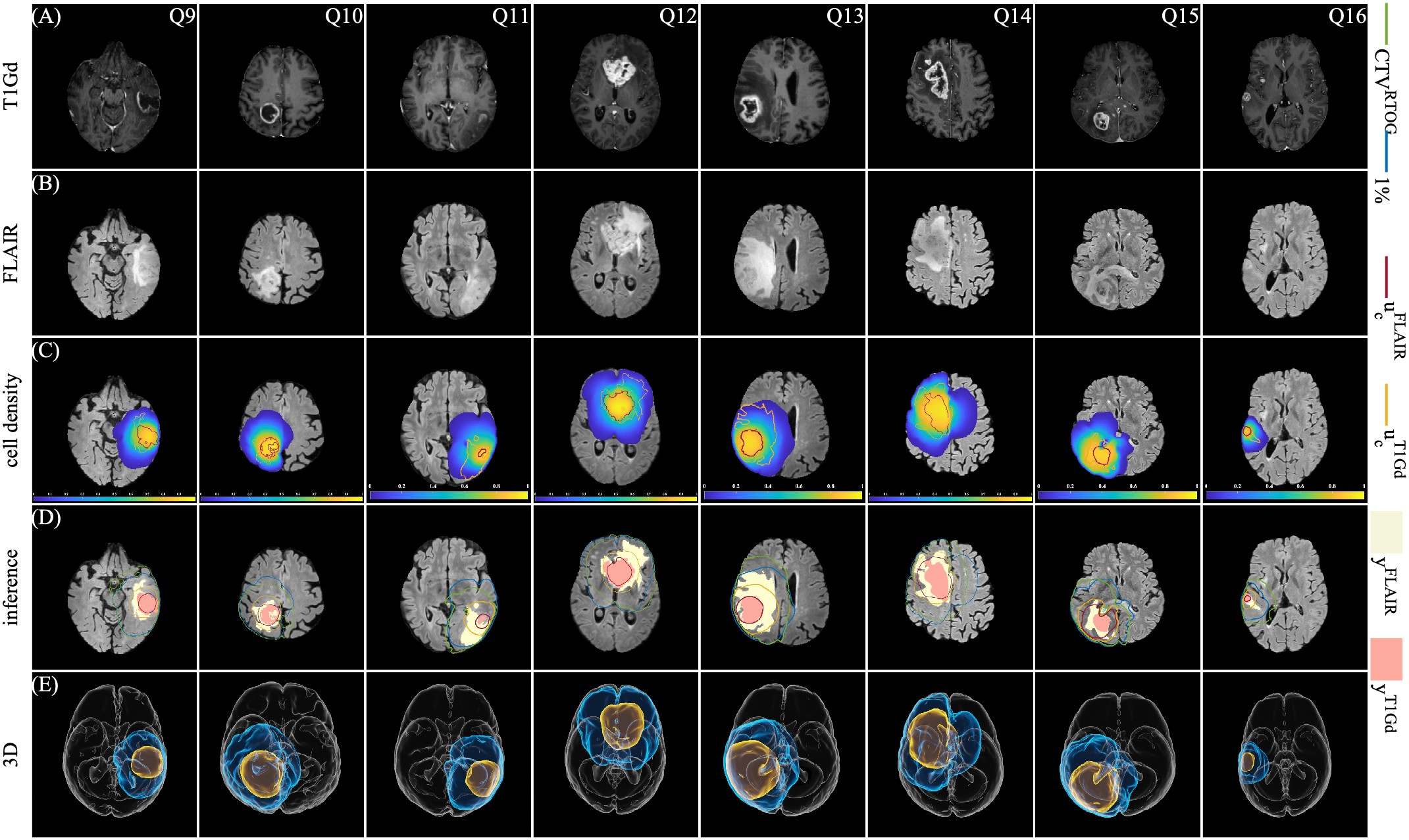}
  \caption{Parameter Estimation Results for Patients Q9-Q16 using \segonly{}. Legends are the same as \ref{f:patseg}. Results using \petseg{} are shown in \ref{ap:patient} Fig.~\ref{f:addpetseg2}.}
  \label{f:addseg2}
\end{figure*}

%%%%%%%%%%%%%%%%%%%%%%%%%%%%%%
% \subsection{Additional Patient Dataset}\label{ss:newpatient}
We consider another set of 16 patients Q1-Q16 and show the $\ufdm$ results from parameter estimation using \segonly{} in Figs.~\ref{f:addseg1} and \ref{f:addseg2}. 
The estimated parameters and metrics are tabulated in \ref{ap:patient}, Table~\ref{t:newpatient}. 
The results are consistent with patients P1-P8: estimations using \petseg{} data are very similar to those using \segonly{} data.
In Table~\ref{t:allpat}, we compare the DICE scores and correlations with FET signals for all patients (24 in total) using characteristic parameters and parameters obtained by fine-tuning with \segonly{} with \petseg{} data. 
Though the characteristic parameters yield good results, fine-tuning leads to better results, while the difference between using \petseg{} and \segonly{} data is small. 
Additional results from the patient datasets are presented in \ref{ap:patient}.
%In Figs.~\ref{f:addpetseg1} and \ref{f:addpetseg2}, we show the results using \petseg{} data for patients Q1-Q16.
%In Table~\ref{t:allpatpinn}, we also compare the DICE scores and correlation with FET signals for all patients using $\upinn$. 
%The $\upinn$ metrics tends to be better than $\ufdm$ metrics. 
%However, as discussed earlier, this can be attributed to the sacrificing the fidelity to the PDE for fitting the data.

\begin{figure}[!h]
  \centering
  \includegraphics[keepaspectratio,width=\linewidth]{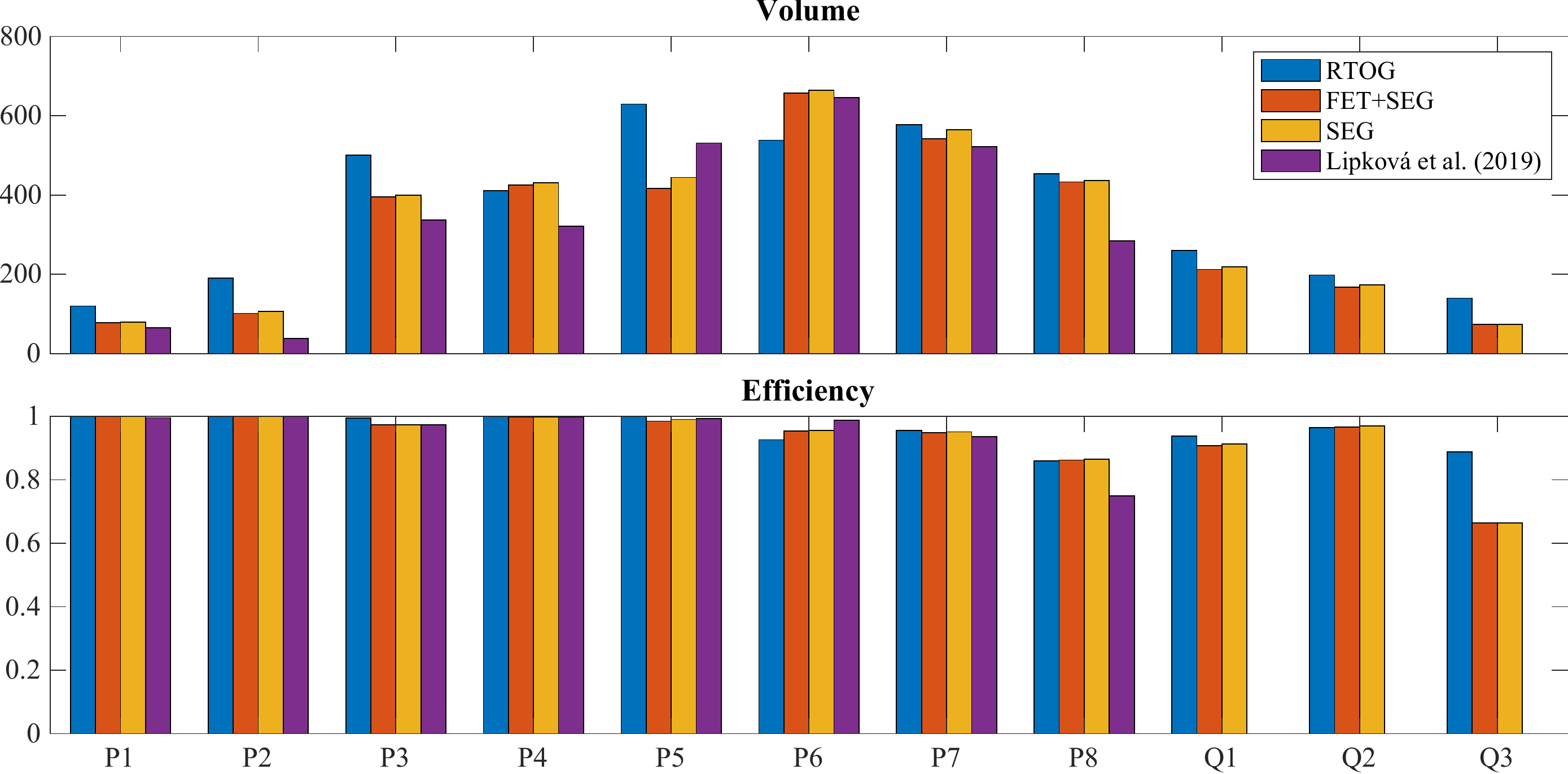}
  \caption{
    Comparison of RTOG CTV (blue) with Personalized CTV (1\% isosurface of predicted tumor cell density) from different parameter inferences: \jana{} (purple), \petseg{} (red), and \segonly{} (yellow).
    (A) Total irradiated volume.
    (B) Efficiency, a percentage tumor core and edema at the first detected recurrence that is covered by the CTV (see text).
    In general, the personalized CTV from \jana{}, \petseg{}, and \segonly{} present reduced or similar irradiation volumes while maintaining efficiency comparable to RTOG CTV.}
    \label{f:voleff}
\end{figure}

%%%%%%%%%%%%%%%%%%%%%%%%%%%%%% CTV
For a quantitative evaluation of the personalized CTVs, we follow \citet{lipkovaPersonalizedRadiotherapyDesign2019} and use the CTV volume and efficiency of the CTV in capturing tumor recurrence as two criteria. A smaller CTV means that less healthy brain tissue is irradiated.
The CTV efficiency is defined as the percentage of the recurrent tumor covered by the CTV:
\begin{equation}
  \eta^{CTV} = {|V^{REC}\cap V^{CTV}|}/{|V^{REC}|},
  \label{eq:eff}
\end{equation}
where $V^{CTV}$ is the CTV region,
$V^{REC}$ is the FLAIR segmentation (tumor core and edema region) of the recurrent tumor and $|\cdot |$ denotes the volume. 
Figure~\ref{f:voleff} compares the volumes and efficiencies of RTOG CTV (blue) with the $\ctvp$ (1\% isosurface of predicted tumor cell density $\ufdm$) using different parameter inference results, \petseg{} (red), \segonly{} (yellow),  from patient data P1-P8 and Q1-Q3 for which we have both T1Gd and FLAIR images at recurrence.
For patients P1-P8, we also include the personalized CTVs based on the maximum a posteriori (MAP) estimates from \citet{lipkovaPersonalizedRadiotherapyDesign2019} (purple). 

The $\ctvp$ volumes and efficiencies using \petseg{} and \segonly{} data from our PINN-based inference are very similar and except for  patients P2 and P8, where the volumes of the $\ctvp$ from \citet{lipkovaPersonalizedRadiotherapyDesign2019} are significantly smaller, the volumes and efficiencies using our PINN-based approach are similar to those from \citet{lipkovaPersonalizedRadiotherapyDesign2019}, which is much more expensive to compute. Compared to $\ctvrtog$, the $\ctvp$ from our study and from \citet{lipkovaPersonalizedRadiotherapyDesign2019} exhibit reduced or similar volumes while maintaining efficiencies comparable to $\ctvrtog$.  Patient Q13 is an outlier, as the recurrent tumor has an exceptionally large edema region, leading to low efficiency for all CTVs. We discuss possible explanations for this in \ref{ap:patient} and Fig.~\ref{f:explainq3}.

\begin{table}[h!]
  \centering
  \begin{tabular}{lccc}
    \toprule
    & {DICE\textsuperscript{T1Gd}} & {DICE\textsuperscript{FLAIR}} & {corr\textsuperscript{FET}} \\
    \hline
    Char & \(0.586 (\pm 0.123)\) & \(0.408 (\pm 0.183)\) & \(0.374 (\pm 0.201)\) \\
    SEG & \(0.695 (\pm 0.091)\) & \(0.685 (\pm 0.160)\) & \(0.416 (\pm 0.224)\) \\
    SEG+FET & \(0.695 (\pm 0.090)\) & \(0.686 (\pm 0.152)\) & \(0.421 (\pm 0.232)\) \\
    \hline
  \end{tabular}
  \caption{Average and standard deviation of DICE scores and correlation with FET signal for all 24 patients using characteristic parameters, or parameters obtained by fine-tuning with \segonly{} with \petseg{} data. Fine-Tuning leads to better results.
  }
  \label{t:allpat}
\end{table}

%%%%%%%%%%%%%%%%%%%%%%%%%%%%%%
\subsection{Discussion}\label{ss:discussion}

%% comparison with Bayesian
We briefly discuss the merits and limitations of our PINN approach, compared with other methods.
% between our PINN approach with Baysesian approach \citep{lipkovaPersonalizedRadiotherapyDesign2019}.
One of the strengths of Bayesian methods is their ability to quantify uncertainties, 
while PINNs provide deterministic outputs based on the minimizer of the loss function and do not inherently provide uncertainty estimates, 
although some methods have been proposed to integrate uncertainty quantification into PINNs \citep{yangBPINNsBayesianPhysicsinformed2021}.
However, Bayesian methods are computationally expensive as they require solving the forward PDE problem many times to estimate the posterior distribution of the parameters, which might takes up to days \citep{lipkovaPersonalizedRadiotherapyDesign2019,ezhovLearnMorphInferNewWay2023}.
Our PINN implementation on TensorFLow \citep{abadiTensorFlowLargescaleMachine2015} takes about 30 min for the pre-training step and 5 min for the fine-tuning step on a single GPU (Nvidia Quadro RTX 8000).
% In a clinical setting, since usually a single prediction ``best'' prediction is often used for treatment planning. 
% In Bayesian methods, usually the maximum a posteriori (MAP) estimate is used.
%% exploration and exploitation
% Since PINN primarily focus on minimizing the loss function, it tends to identify parameters that are close to the initial guess.
% In our experiments, the initial location of the tumor barely move away from the initial guess, which is the centroid of the T1Gd segmentation. In contrast, Bayesian methods can explore a larger region of the parameter space. 
%% computational cost
% The computational cost strongly depends on the implementation and the hardware,

Conventional PDE-constrained optimization methods, such as the adjoint method, are very efficient and solve the PDE using numerical solvers, which come with well-understood error estimators, as well as convergence and stability guarantees
\citep{hogeaImagedrivenParameterEstimation2008,gholamiInverseProblemFormulation2016,subramanianWhereDidTumor2020,subramanianMultiatlasCalibrationBiophysical2020,subramanianEnsembleInversionBrain2023}. In contrast, the training of PINNs remains an active area of research \citep{karniadakisPhysicsinformedMachineLearning2021,krishnapriyanCharacterizingPossibleFailure2021,wangWhenWhyPINNs2022,haoGaussNewtonMethod2023,chenRandomizedNewtonMethod2022}.
% Therefore, in our method, we estimate the parameters and solve the PDE using FDM, instead of evaluating the PINN.
One notable advantage of PINNs is their flexibility and simplicity in implementation: modifying the model primarily involves changing the loss function.
In contrast, the adjoint method often requires derivation of the adjoint equation, careful regularization, and implementation of the adjoint solver, adding to the technical demands. 

Another flexible machine learning approach, known as ODIL \citep{karnakovOptimizingDIscreteLoss2022}, was recently developed to estimate parameters of the Fisher-KPP PDE using MRI and FET-PET scans for GBM \citep{Balcerak2023}. Like the PINN, this method also minimizes a loss function that contains data and the PDE residual.  In ODIL the residual of the PDE is calculated using a grid-based discretization of the PDE and the variables to be minimized are the solutions at the grid points in 4D (space+time) rather than weights of a neural network as in the PINN. Although the number of grid points can be very large, the multigrid method and the sparsity of the discrete systems can be used to improve efficiency. While promising, the current implementation uses downsampling of the medical images.

% Comparison with ML
Another class of methods aims to learn the mapping between image and parameters directly, either by training a neural network \citep{ezhovLearnMorphInferNewWay2023}, or by posing the inverse problem as a database query task \citep{ezhovForloopAllYou2022}.
These methods require a large number of forward PDE solves to generate a synthetic dataset and  can take days to compute, which is generally not a concern in practice as it is done offline.
The inference is extremely fast since it only requires evaluating the neural network or a for-loop search.
In \citep{ezhovLearnMorphInferNewWay2023}, 100,000 synthetic tumors were generated for training, while the inference was done in about 5 minutes.
% More complicated PDE model can be used without affecting the inference time.
However, this approach does not allow any patient specific fine-tuning, thus might have larger errors if the patient data is significantly different from the training data.

Currently, our method requires pre-training for each patient using patient-specific characteristic parameters, obtained by solving a 1D model in spherically symmetric geometries, and an associated patient-specific characteristic solution generated by solving the Fisher-KPP PDE with characteristic parameters in the patient brain geometry. From Table~\ref{t:allpat}, we can see that the characteristic solution already gives remarkably effective predictions of tumor cell densities, as measured by DICE scores and correlations with FET-PET signals.
This method works well because it incorporates insights into the tumor growth dynamics from image features, and solving the PDE in patient geometries captures the heterogeneity of the spreading.
Fine-tuning with patient data is necessary to obtain more accurate results, but the improvement is also limited, because the Fisher-KPP equation is a simplified model that does not capture the full complexity of GBM growth, such as the mass effect or the presence of necrosis. 

One might wonder if using an averaged set of the characteristic parameters can be used to pre-train a single network for all patients. We tested this (see \ref{ap:fixchar}) and found that such a pre-trained single network does not work well with our current framework because the tumors have different scales, as demonstrated by the wide range of patient-specific parameters.
Without the patient-specific characteristic parameters, the pre-trained network can be far away from the solution, leading to difficulty in fine-tuning.
Nevertheless, the idea of a common pre-trained network for all patients is appealing as it can significantly increase the inference speed. Consequently, in future work we will explore other approaches for developing a single network that can be used for all patients, including transfer learning \citep{xuTransferLearningBased2023,desaiOneShotTransferLearning2021,gaoSVDPINNsTransferLearning2022} and neural operators \citep{liFourierNeuralOperator2020,luLearningNonlinearOperators2021}

% Currently, our method requires a separate training workflow for each patient, we expect to significantly speed-up our algorithms by using approaches from transfer learning \citep{xuTransferLearningBased2023,desaiOneShotTransferLearning2021,gaoSVDPINNsTransferLearning2022} or neural operators \citep{liFourierNeuralOperator2020,luLearningNonlinearOperators2021}, to pre-train a single network for all patients with different $\cD$ and $\cR$. This is currently under study.

\section{Conclusion}\label{s:conclusion}

We presented a method employing Physics-Informed Neural Networks (PINNs) to estimate patient-specific parameters of a reaction-diffusion PDE model for the growth of GBM tumors from a single snapshot of structural MRI scans.
We applied a simple procedure to estimate patient-specific characteristic parameters of the PDE model using a 1D model in spherically symmetric geometries and a grid-search procedure.
The characteristic parameters allowed us to scale the PDE to learn parameters that are close to unity,
and helped us to pre-train the PINN to learn a characteristic solution before fine-tuning the network with patient data to estimate the patient-specific parameters, leading to a more efficient method.
Verified on synthetic and patient datasets, the results presented here provide a proof-of-concept that PINNs hold great promise for combining medical images and biophysical models for patient-specific tumor growth prediction and treatment planning.

% Future direction includes incorporating mass effect or multi-species tumor model that accounts for the necrotic region \citep{subramanianSimulationGlioblastomaGrowth2019}.
% Our framework can also be extended to study the effect of treatment as done in \citep{jiPostRadiotherapyPETImage2022,chaudhuriPredictiveDigitalTwin2023}. 

% \begin{comment}

\section*{Declaration of competing interest}
The authors declare that they have no known competing financial interests or personal relationships that could have appeared to
influence the work reported in this paper

\section*{Data Availability}
The pre-operative MRI scan of patient P1-P8 are available from \citet{lipkovaPersonalizedRadiotherapyDesign2019}.
Other data will be made available on request.
The PINN code is available at \url{https://github.com/Rayzhangzirui/pinn}

\section*{Acknowledgments}
R.Z.Z and J.S.L thank Jana Lipkova, Peter Chang, and Xiaohui Xie for helpful discussions, and Babak Shahbaba for the GPU resources.
J.S.L. acknowledges National Institutes of Health for partial support through grant nos. 1U54CA217378-01A1 for a National Center in Cancer Systems Biology at UC Irvine and P30CA062203 for the Chao Family Comprehensive Cancer Center at UC Irvine. In addition, J.S.L. acknowledges support from DMS-1763272, DMS-1936833, DMS-1953410, DMS-2309800 and the Simons Foundation (594598QN) for an NSF-Simons Center for Multiscale Cell Fate Research. 
BM and MB acknowledge support from the Helmut Horten Foundation.

% clearpage would prevent the appendix figure from floating 
% \clearpage
\appendix

%%%%%%%%%%%%%%%%%%%%%%%%%%%%%%
\section{Training of PINN} \label{ap:training}

We use a fully connected neural network with 4 hidden layers. Each layer has 128 neurons, and the  tanh activation function is used. 
Different training strategies are employed for pre-training and fine-tuning:
Pre-training is done using the characteristic solution $\bufdm$ as data. 
We employed the ADAM optimizer with initial learning rate of 0.001, followed by L-BFGS-B optimizer with maximum number of variable metric corrections of 100, maximum number of line search steps of 50, and a tolerance for termination of 2.3e-09.
We take $\Nres = 20,000$.
The loss is also computed on the same number of testing points. The maximum number of iterations is 20,000, with early stopping if the training loss or testing loss does not decrease for 1000 iterations.
For fine-tuning, we use the same optimization procedure (ADAM followed by L-BFGS-B),
but with a smaller dataset $\Nres = \Ndat = 5,000$, a smaller maximum iteration of 5000 steps, and a 
smaller initial learning rate of 0.0001. 

\begin{figure}[!t]
  \centering
  \includegraphics[keepaspectratio,width=\linewidth]{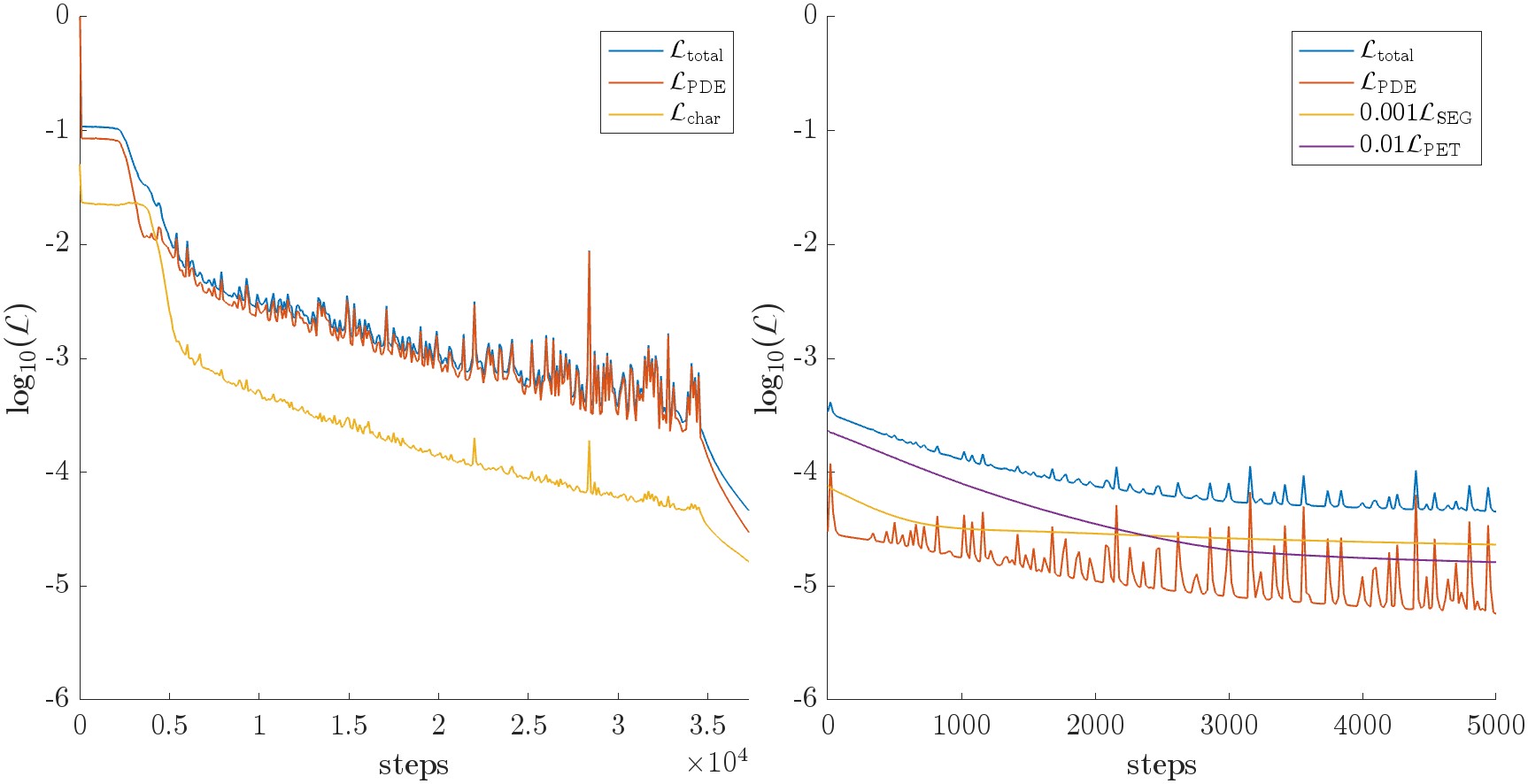}
  
  \caption{
    Example of training loss for patient P5 with \petseg{}.
    The left figure shows the training loss for pre-training with residual loss and $\ufdm$.
    The right figure shows the training loss for fine-tuning with segmentation loss and PET loss.
    The weights are chosen such that the two loss magnitudes are similar. 
    }
    \label{f:loss}
\end{figure}

In Fig.~\ref{f:loss}, we show an example of the training loss for patient P5 with \petseg{}. 
The left figure shows the training loss for pre-training with residual loss and $\ufdm$.
The right figure shows the training loss for fine-tuning with segmentation loss and PET loss.
The $\lseg$ is weighted by 0.001 and $\lfet$ (if necessary) is weighted by 0.01, while all the other losses are weighted by 1.
These weights are chosen such that the magnitude of the losses are comparable, and are constant throughout the training.
While we experimented with some adaptive weighting schemes, which balance different loss terms according to coefficients of variation \citep{groenendijkMultiLossWeightingCoefficient2020}, by the norm of the gradients \citep{chenGradNormGradientNormalization2018,wangUnderstandingMitigatingGradient2021}, or the standard deviation of the gradients \citep{madduInverseDirichletWeighting2022}, 
using the fixed weights listed above achieved the best results across all synthetic and patient datasets, without significantly increasing the training time.

%We find that these weights achieve good results across all synthetic and patient datasets. 
% We mentioned that we experiment with some adaptive weighting scheme, but did not find successful results.

%%%%%%%%%%%%%%%%%%%%%%%%%%%%%%
\section{Distribution of collocation points} 
\label{ap:xdist}

We explain how we sample the collocation points for the residual loss, 
which affects the training efficiency and accuracy.
In \citet{luDeepXDEDeepLearning2021}, an adaptive residual refinement technique is used to adaptively increase the density of collocation points in the region where the residual is large. 
In \citet{wangRespectingCausalityAll2022}, the residuals at early times are adaptively given higher weights.
We adopt a similar approach in our work here.
We know that the solution of the PDE will have larger gradients at early times, and the solution will be concentrated at the center of the tumor. 

\begin{figure}[!t]
  \centering
  \includegraphics[keepaspectratio,width=\linewidth]{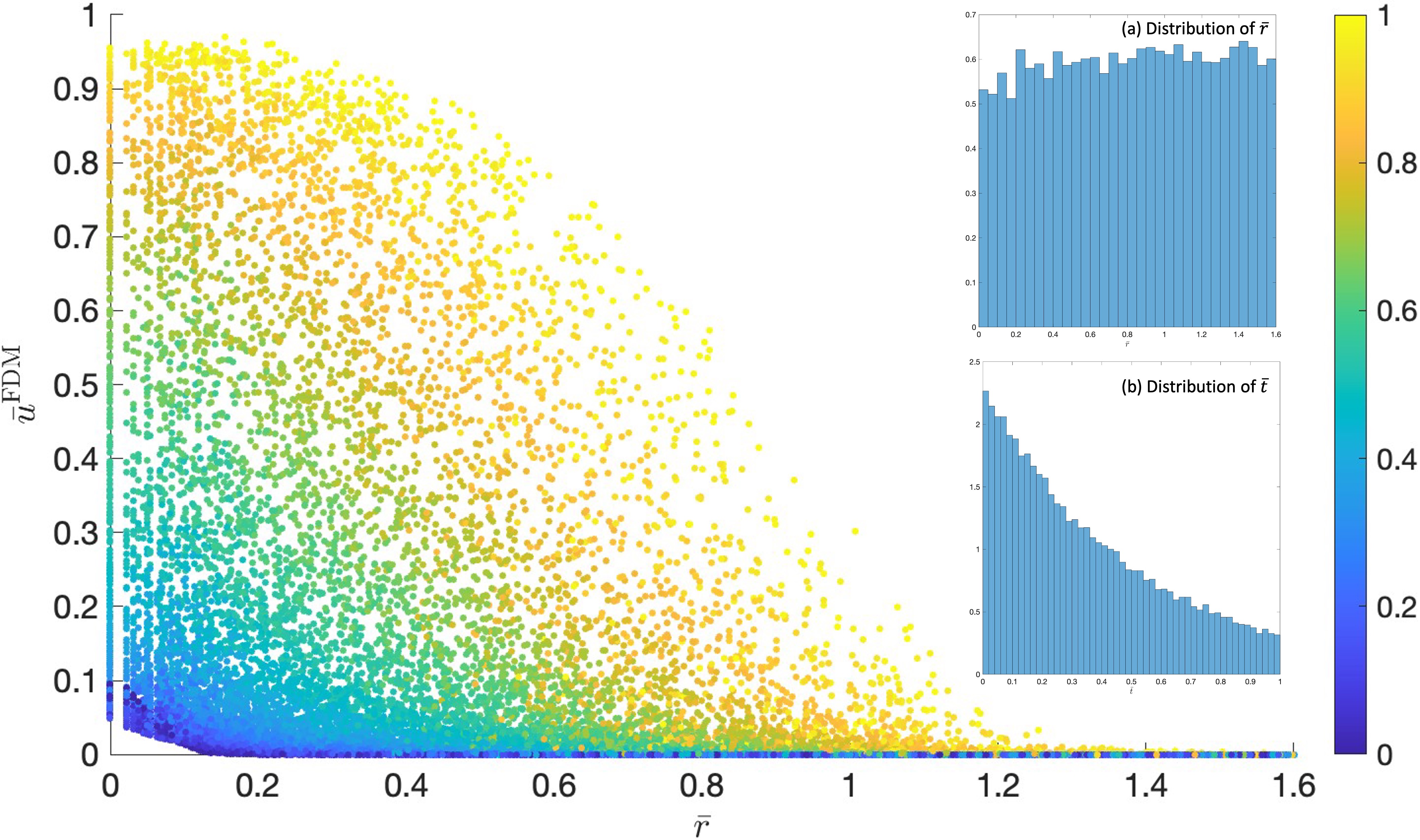}  
  \caption{The distribution of collocation points for the residual loss in the radial direction is shown as a scatter plot with $\bufdm$ and r (radii of $\xres$) and color-coded by time spanning from 0 to 1. 
  The PDE solution is concentrated at the center where there is a larger gradient at early times, and the collocation points are densely sampled at the center and at early times. 
  Inset (a) shows a histogram of r (normalized to a probability density function), highlighting that due to near-uniform distribution in polar coordinates, $\xres$ is denser at the center. 
  Inset (b) shows a histogram of time (normalized as a probability density function), indicating a denser sampling in the earlier time periods.}
  \label{f:colloc}
\end{figure}

The distribution of the collocation points is illustrated in Fig.~\ref{f:colloc}.
Let $\icx$ be the centroid of $\segt$, choose the radius R such that the spherical ball $B$ with center $\icx$ and radius R encloses the region where $\bar{u}^{FDM}>0.001$.
Next, we sample the grid points $\vb x_i$ in $C$ such that $\vb x_i$ are dense in the center, sparse near the boundary, and $t_i$ are dense at early times and sparse at late times, because the solution $u$ has larger gradients in the tumor center and at early times.
More specifically, $t_i$ is sampled from a truncated exponential probability density function $\lambda e^{-\lambda t}$, with $\lambda = 0.5$ and $t \in [0, 1]$. 
Each spatial coordinate $\vb x_i$ is resampled with weight $1/(r_i^2)$ in 3d, where $r_i$ is the distance from $\vb x_i$ to the tumor center $\icx$. 
% $\epsilon$ is a small number to avoid singularity, and is chosen such that $r_i$ is almost uniform.
% We do not adaptively modify the distribution during training as that could slow down the training.
We experimented with other static distributions of the collocation points, including uniformly sampling the temporal and spatial coordinates, or densely sampling the spatial coordinates at the transition region between white and gray matter (where the diffusion coefficient changes sharply),
and an adaptive residual refinement technique following \citep{luDeepXDEDeepLearning2021}, which recruits the 100 testing points with the largest residual loss into the training points every 1000 iterations,
but we did not find significant improvement in accuracy.

%%%%%%%%%%%%%%%%%%%%%%%%%%%%%%%%%%%%%%%%%%%%%%%%%%%%%%%%%%%%
\section{Additional results using synthetic data}
\label{ap:synthetic}

\begin{figure}[!t]
  \centering
  \includegraphics[keepaspectratio,width=0.5\columnwidth]{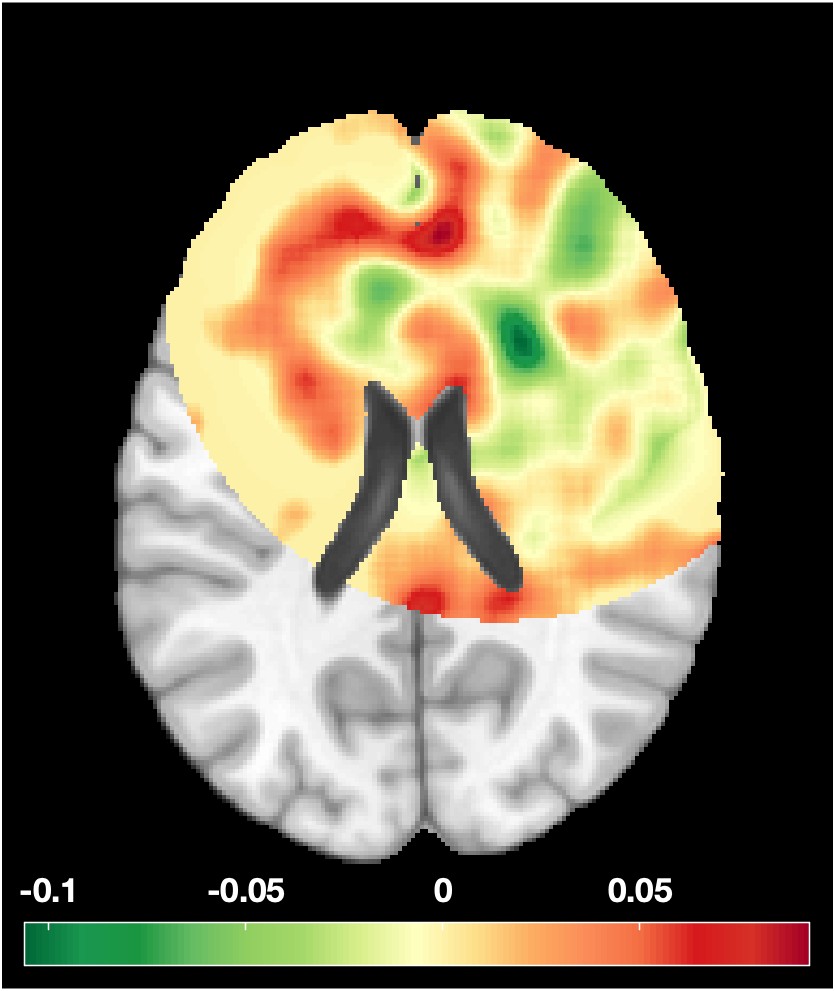}
  \caption{Illustration of spatially correlated noise, as employed in Fig.~\ref{f:patdetail}. Noise is produced by sampling from a Gaussian distribution (mean 0 and standard deviation 2) at each pixel and subsequently applying a Gaussian filter of size 13.}
  \label{f:noise}
\end{figure}

For all the synthetic data S1-S8, after we obtain the ground truth cell density $\ufdm$ by solving the Fisher-KPP PDE using the FDM,
we generate spatially correlated noise $v$ by sampling from a Gaussian distribution (mean 0 and standard deviation 2) at each pixel and subsequently apply a Gaussian filter of size 13, as shown in Fig.~\ref{f:noise}.
We add $v$ to $u$ and threshold the result in [0, 1]. The synthetic segmentation and FET signal are generated based on this noisy cell density.

In Table~\ref{t:syn}, we show the estimated parameters for the synthetic dataset S1-S8, comparing the results using $\petseg$ and $\segonly$ data with the ground truth parameters.
In Supplementary Materials, we show all S1-S8 results as in Fig.~\ref{f:syndetail}.

\pgfplotstableread[col sep=comma]{tab_syn.csv}\mydata
\begin{table*}[!t]
  \centering
  \pgfplotstabletypeset[
    row sep=crcr,
    string type,
    header=false,
    col sep=comma,
    every last row/.style={after row=\hline},
    every first row/.style={before row=\hline},
    every head row/.style={output empty row},
    every nth row={3[+1]}{before row=\hline},  % insert hline every 3 rows
    ]\mydata
    \caption{Estimated parameters from the synthetic dataset using \segonly{} and \petseg{}, compared with ground truth (GT) values.}
    \label{t:syn}
\end{table*}

%%%%%%%%%%%%%%%%%%%%%%%%%%%%%%
\section{Additional results for patient tumors}
\label{ap:patient}

\begin{figure}[!t]
  \centering
  \includegraphics[keepaspectratio,width=0.5\textwidth]{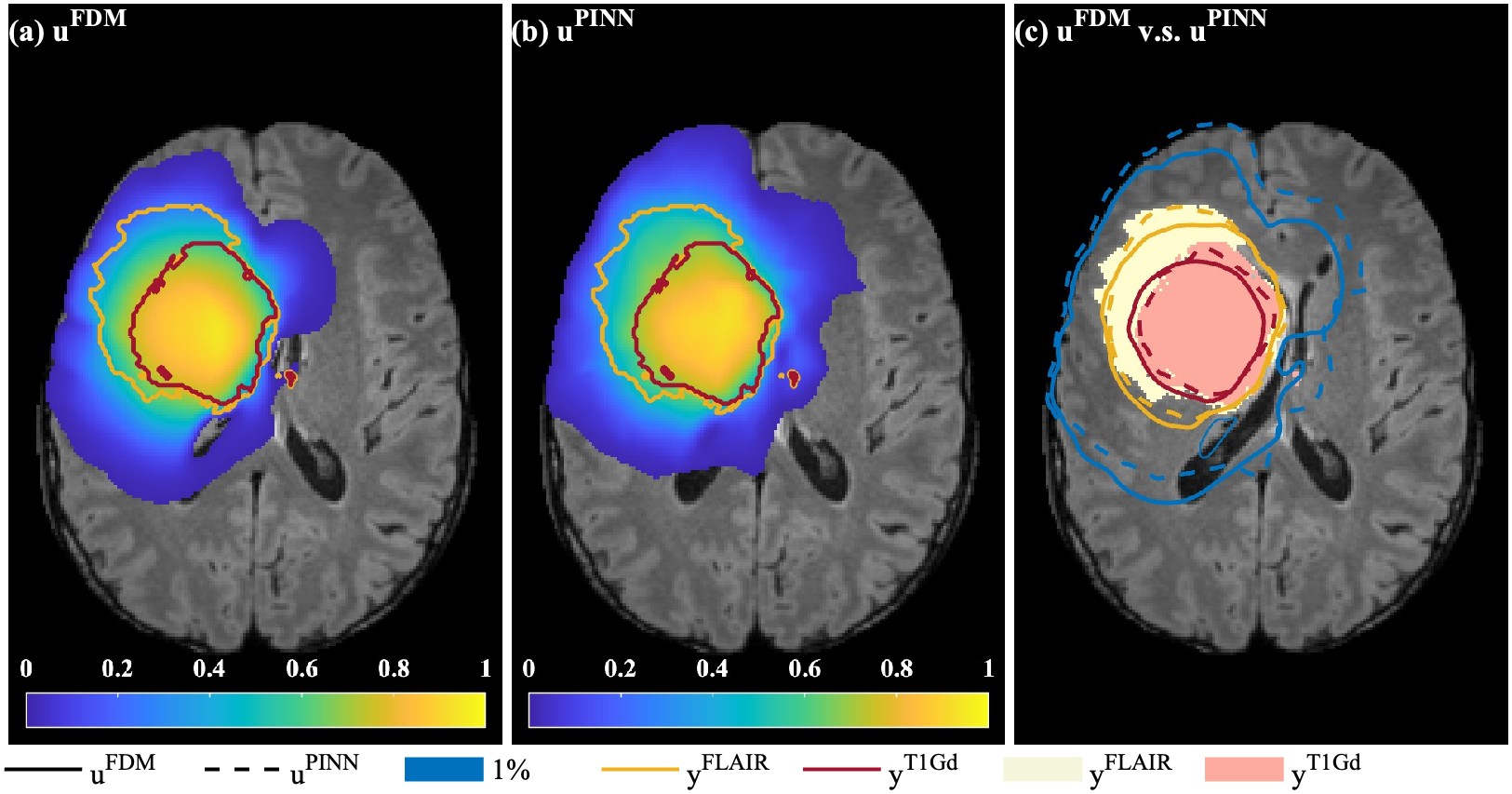}
  \caption{Comparison of predicted tumor cell densities $\ufdm$ and $\upinn$ for patient P5 using \segonly{} data, as supplement to Fig.~\ref{f:patdetail}. $\upinn$ is influenced by the noise.
  }
  \label{f:patpinnvsfdm}
\end{figure}

In Fig.~\ref{f:patpinnvsfdm}, as supplement to Fig.~\ref{f:patdetail}, we compare $\ufdm$ and $\upinn$ for patient P5 using \segonly{} data.
Similar to the results in the synthetic case (Fig.~\ref{f:syndetail}), the predicted segmentations of $\upinn$ align more closely with the actual segmentations than $\ufdm$ does, while the 1\% contour and the predicted density is noisier than that from $\ufdm$.
This suggests that when the model cannot fully account for the data, the PINN reduces the overall cost by reducing the data loss at the expense of the PDE loss. Accordingly, the  $\upinn$ metrics, including DICE scores and correlations with FET signal, tend to be better than $\ufdm$ metrics (see \ref{ap:patient},  Table~\ref{t:allpatpinn}).
This might not be desired in terms of PDE-constrained optimization, as the constraint (e.g., the PDE) is not satisfied.
However, in a data-driven approach, the PDE loss can be considered as a regularization term that prevents overfitting \citep{Balcerak2023}.

\begin{table}[h!]
  \centering
  \begin{tabular}{lccc}
    \toprule
    & {DICE\textsuperscript{T1Gd}} & {DICE\textsuperscript{FLAIR}} & {corr\textsuperscript{FET}} \\
    \hline
    SEG & \(0.747 (\pm 0.089)\) & \(0.678 (\pm 0.162)\) & \(0.421 (\pm 0.231)\) \\
    SEG+FET & \(0.753 (\pm 0.084)\) & \(0.690 (\pm 0.165)\) & \(0.473 (\pm 0.217)\) \\
    \hline
  \end{tabular}
  \caption{Average and standard deviation of DICE scores and correlation with FET signals for all 24 patients using $\upinn$. 
  The metrics are in general better than those using $\ufdm$ (in Table \ref{t:allpat}) because $\upinn$ might sacrifice fidelity to the PDE to fit the data.
  }
  \label{t:allpatpinn}
\end{table}

\begin{figure*}[!t]
  \centering
  \includegraphics[keepaspectratio,width=\textwidth,height=0.9\textheight]{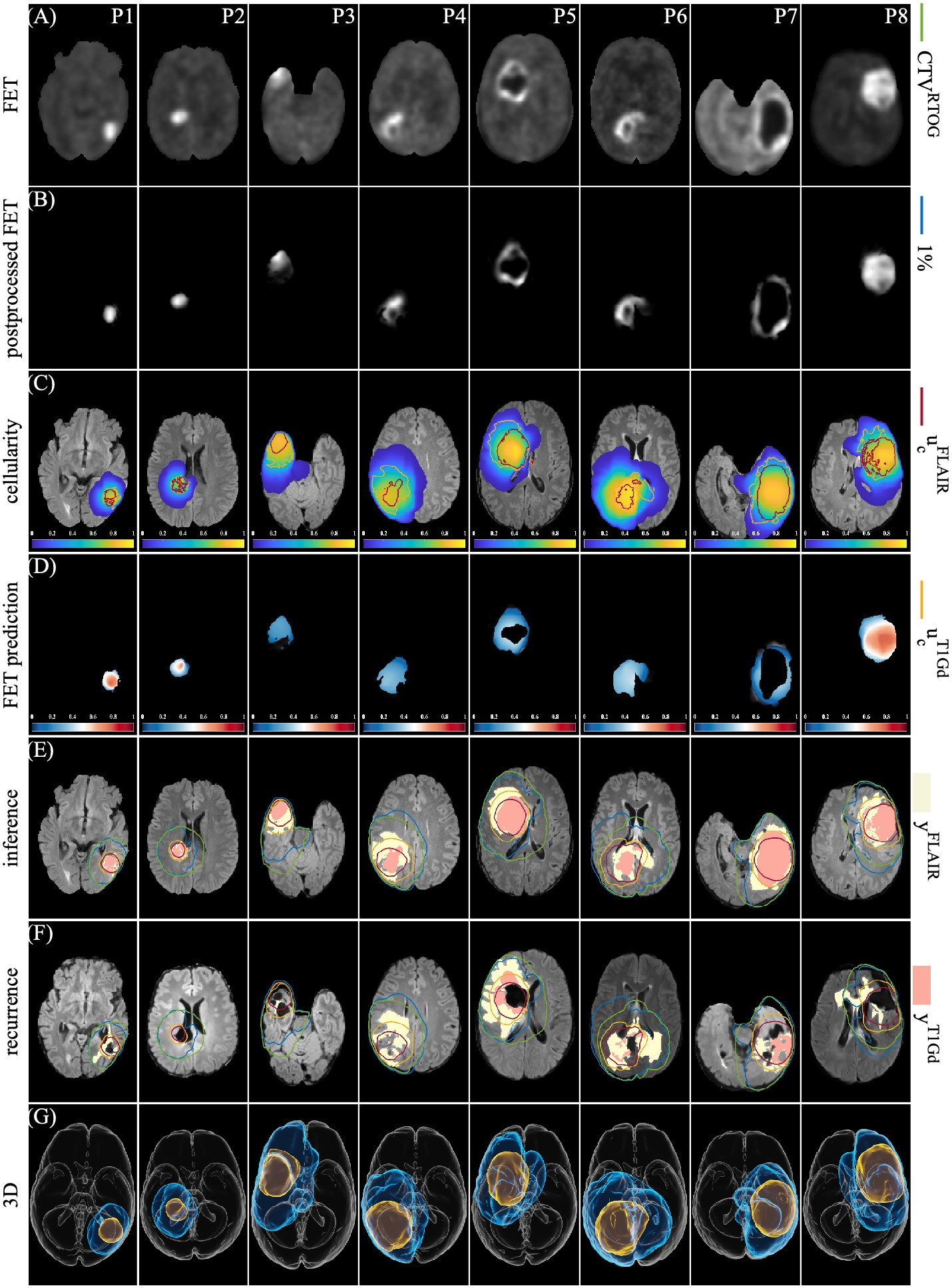}
  \caption{Parameter Estimation Results for Patients P1-P8 using \petseg{}.
  Pre-operative T1Gd and FLAIR are identical to Fig.~\ref{f:patseg}.
  (A) FET signal.
  (B) Pre-processed PET by subtracting the background average and normalized to [0,1].
  (C) Superimposition of predicted tumor cell density $\ufdm$ on actual segmentations: T1Gd (yellow line) and FLAIR (red line).
  (D) Predicted FET signal.
  (E) Overlap of inferred segmentations: $\ucf$ (yellow line) and $\uct$ (red line) on actual segmentations: T1Gd (beige fill) and FLAIR (pink fill); Margins for $\ctvrtog$ (green) and $\ctvp$ (blue).
  (F) Tumor recurrence, compared with with margins for $\ctvrtog$ (green) and $\ctvp$ (blue).
  (G) 3D reconstructions of $\ctvp$ and $\ucf$ isosurfaces.}
  \label{f:patsegpet}
\end{figure*}

We show the results for patients P1-P8 using \petseg{} data in Fig.\ref{f:patsegpet}.
In Table~\ref{t:patparammetric}, the estimated parameters P1-P8 are shown, using both \segonly{} and \petseg{} data, and these are compared with the results from \citet{lipkovaPersonalizedRadiotherapyDesign2019}.
We also compute various metrics to evaluate the performance of the methods, including DICE scores for both FLAIR and T1Gd segmentations, denoted as $DICE^{FLAIR}$ and $DICE^{T1Gd}$, as well as correlations with FET signals, denoted as \corfet{}.
For results using \petseg{} data, \corfet{} is the correlation between the predicted FET signal and the predicted FET (Eq~\eqref{eq:fet}).
For results using \segonly{} data, which does not consider the imaging model for FET signal,
\corfet{} is the correlation between the predicted FET signal and the predicted cell density.
Note that the imaging model of FET in \jana{} does not include the intercept A in Eq. (\ref{eq:fet}).

We compare the outcomes using \petseg{} data and using \segonly{} alone. 
It is observed that the estimated parameters and metrics are quite similar; 
however, incorporating the PET data loss appears to slightly enhance the correlation with the FET signal,  at the expense of a marginally lower DICE score for the segmentations.

We also compare the parameters estimated from \citet{lipkovaPersonalizedRadiotherapyDesign2019} with those obtained through our method.
For parameters in \citet{lipkovaPersonalizedRadiotherapyDesign2019}, the MAP estimate of $(D, \rho,\tend)$ is used to compute $D/\rho$ and the non-dimensional $\cD$ and $\cR$ based on our estimated $\cD$ and $\cR$.
We can see that $\cR$ are similar, whereas there is a larger discrepancy in $\cD$ values for P2, P3, P5, and P7.
The $D/\rho$ ratio exhibits a larger discrepancy for P2 and P3.
In terms of the metrics, the performance is similar.
The observed differences could be ascribed to the Bayesian methods' capability to explore a larger parameter space, whereas our training process tends to find parameters close to the initial guess.
Despite the variation in parameters, the DICE score and correlation with the FET signal remain similar.
In \jana{}, the DICE score of T1Gd segmentation is 0, potentially due to a different definition of $\segt$: in their method, it only comprises the enhancing region, while ours additionally includes the necrotic core.
Given there is little T1Gd enhancement for this patient, the majority of the inference is dictated by their model’s attempt to fit the fact that cell density below the cut-off is not discernible outside of the T1Gd-enhancing region, resulting in a high $\uct$.

\begin{figure*}[!h]
  \centering
  \includegraphics[keepaspectratio,width=\textwidth]{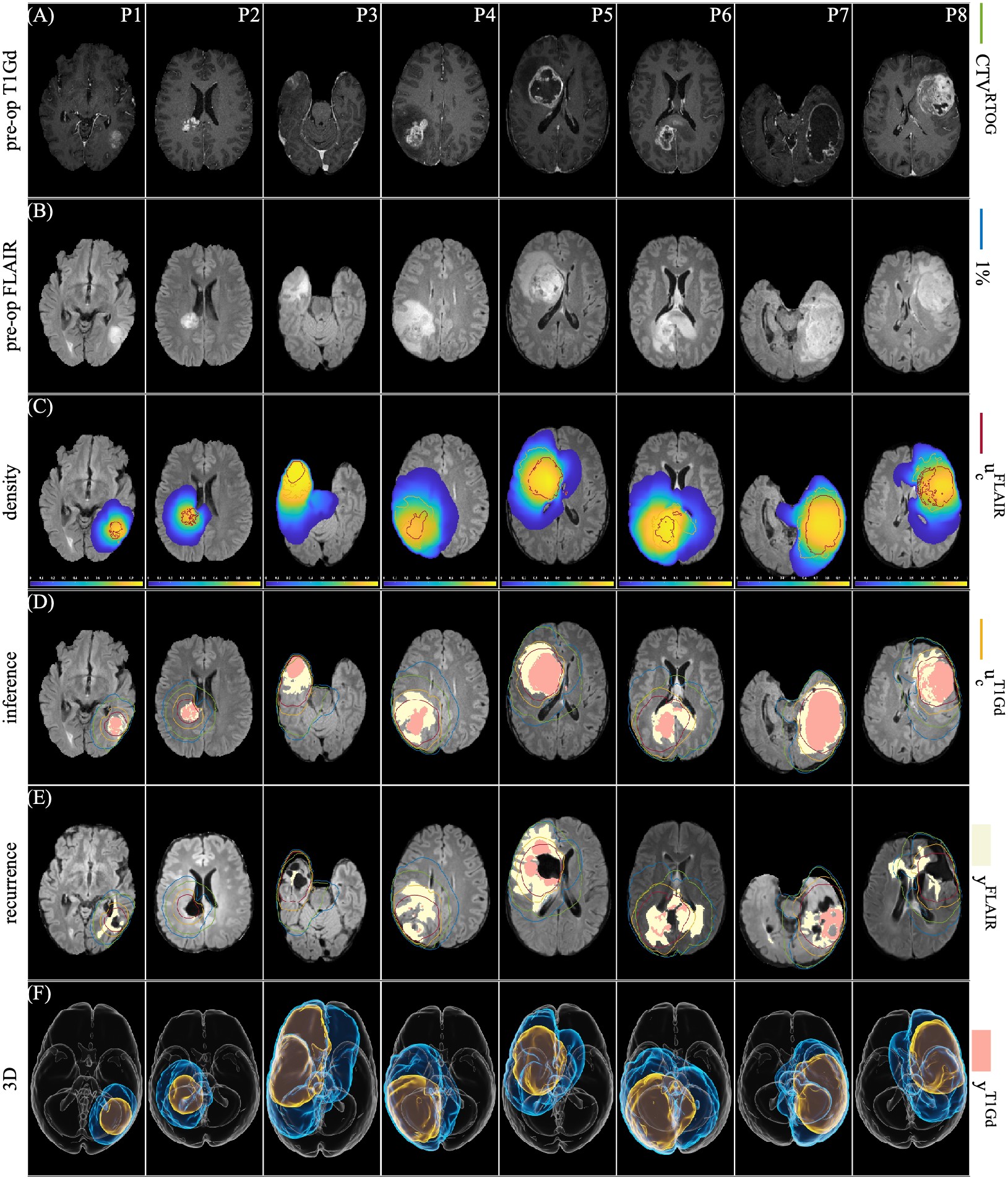}
  \caption{Results for Patients P1-P8 using characteristic parameters without fine-tuning. For all cases, $\muD = \muR = 1$, $m=1$, $A=0$, $\uct = 0.6$, $\ucf = 0.35$.
  (a) Pre-operative T1Gd.
  (b) Pre-operative FLAIR.
  (c) Superimposition of predicted tumor cell density $\ufdm$ on actual segmentations: T1Gd (yellow line) and FLAIR (red line).
  (d) Predicted FET signal.
  (d) Overlap of inferred segmentations: $\ucf$ (yellow line) and $\uct$ (red line) on actual segmentations: T1Gd (beige fill) and FLAIR (pink fill); Margins for $\ctvrtog$ (green) and $\ctvp$ (glue, 1\% contour of $\ufdm$).
  (e) Tumor recurrence, compared with with margins for $\ctvrtog$ (green) and $\ctvp$ (blue).
  (f) 3D reconstructions of $\ctvp$ and $\ucf$ isosurfaces.
  }
  \label{f:patchar}
\end{figure*}

In Table~\ref{t:charmetric}, we show the metrics for predictions using the characteristic parameters, that is, based on $\bufdm$. 
For all cases, 
$\muD = \muR = 1$, $m=1$, $A=0$, $\uct = 0.6$, $\ucf = 0.35$. 
{The morphologies of these patient-specific characteristic tumors are shown in Fig.~\ref{f:patchar}}.
Thus, although it is very efficient to generate the characteristic tumors, without fine-tuning, the performance is not as good as the results with fine-tuning (compare with Table.~\ref{t:patparammetric} and Fig. \ref{f:patseg})
% All the solve the PDE using our simple FDM solver, which is less accurate different from the multiresolution FDM solver used in \citep{lipkovaPersonalizedRadiotherapyDesign2019}.

\pgfplotstableread[col sep=comma]{tab_pat_compare.csv}\mydata
  \begin{table*}[!t]
    \centering
    \pgfplotstabletypeset[
      string type,
      header=false,
      col sep=comma,
      every last row/.style={after row=\hline},
      every first row/.style={before row=\hline},
      every head row/.style={output empty row},
      every nth row={3[+1]}{before row=\hline},  % insert hline every 3 rows
      ]\mydata
      \caption{Parameters estimated for Patients P1-P8 using \petseg{} and \segonly{} data, compared with results from \jana{}($^*$). 
      Parameters that are not included in the model are marked with ``-''.
      Metrics include DICE score for both FLAIR and T1Gd segmentations, as well as correlation with the FET signal (see \ref{ap:patient}).
      }
      \label{t:patparammetric}
  \end{table*}

  \pgfplotstableread[col sep=comma]{tab_pat_char_metric.csv}\mydata
  \begin{table*}[!t]
    \centering
    \pgfplotstabletypeset[
      string type,
      header=false,
      col sep=comma,
      every last row/.style={after row=\hline},
      every first row/.style={before row=\hline},
      every head row/.style={output empty row},
      every nth row={1}{before row=\hline}, 
      ]\mydata
      \caption{Metrics for results using the characteristic parameters without fine-tuning. For all cases, $\muD = \muR = 1$, $m=1$, $A=0$, $\uct = 0.6$, $\ucf = 0.35$. Without fine-tuning, the metrics are not as good as the results in Table.~\ref{t:patparammetric}.
      }
      \label{t:charmetric}
  \end{table*}

\begin{figure*}[!h]
  \centering
  \includegraphics[keepaspectratio,width=\textwidth]{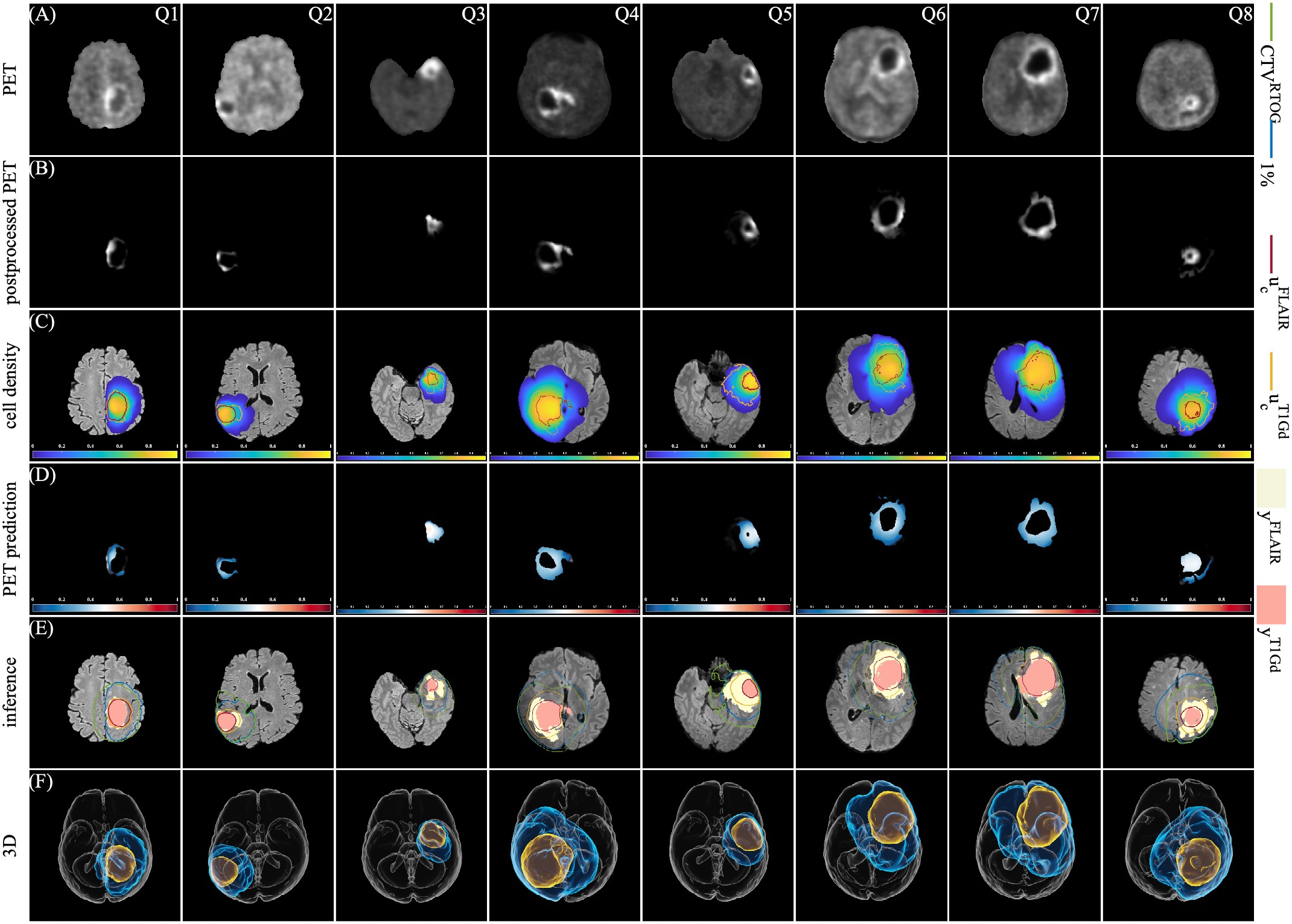}
  \caption{Parameter Estimation Results for Patients Q1-Q8 using \petseg{}. Results with \segonly{} is shown in Fig.~\ref{f:addpetseg1}.
  }
  \label{f:addpetseg1}
\end{figure*}

\begin{figure*}[!h]
  \centering
  \includegraphics[keepaspectratio,width=\textwidth]{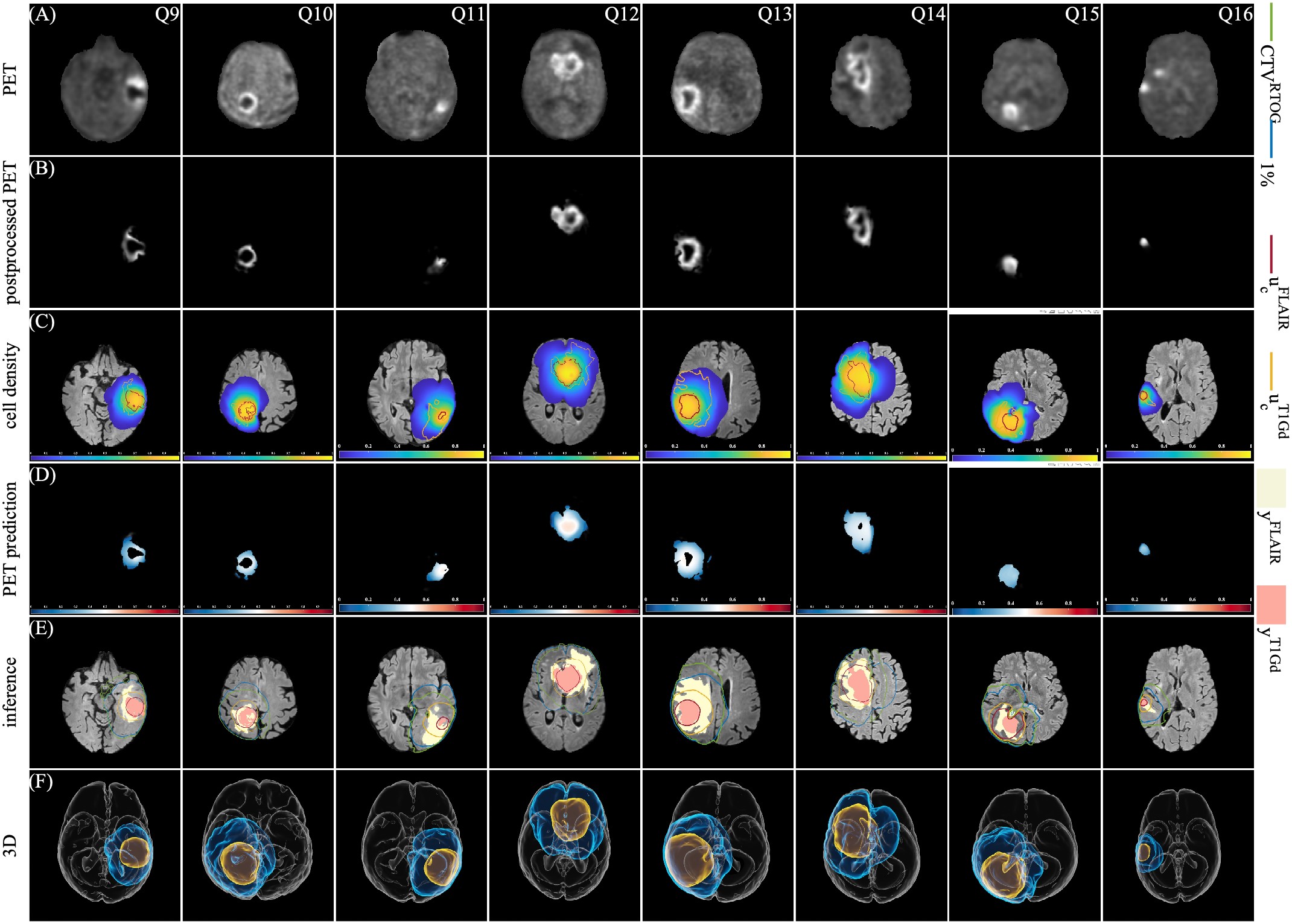}
  \caption{Parameter Estimation Results for Patients Q9-Q16 using \petseg{}. Results with \segonly{} is shown in Fig.~\ref{f:addpetseg1}.
  }
  \label{f:addpetseg2}
\end{figure*}

\begin{figure}[!h]
  \centering
  \includegraphics[keepaspectratio,width=\linewidth]{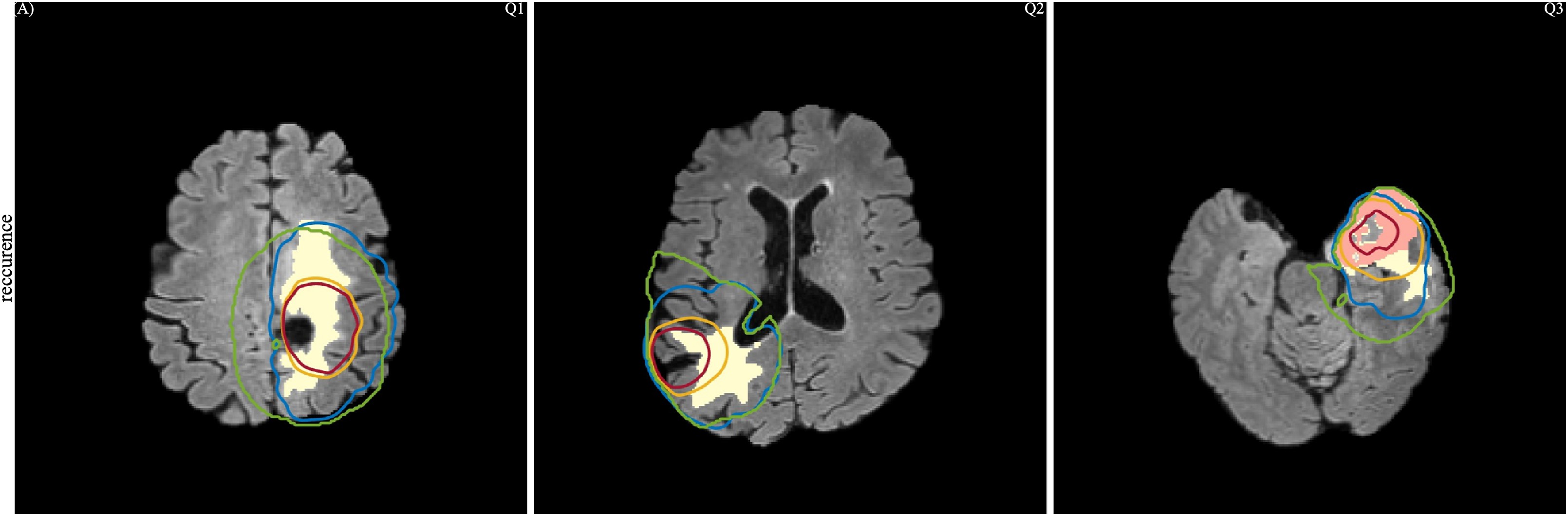}
  \caption{Recurrent tumors from patients Q1-Q3 and PDE model predictions using \segonly{}, as supplement to Fig.~\ref{f:addseg1}.}.
  \label{f:benerec}
\end{figure}

\begin{figure}[!h]
  \centering
  \includegraphics[keepaspectratio,width=\linewidth]{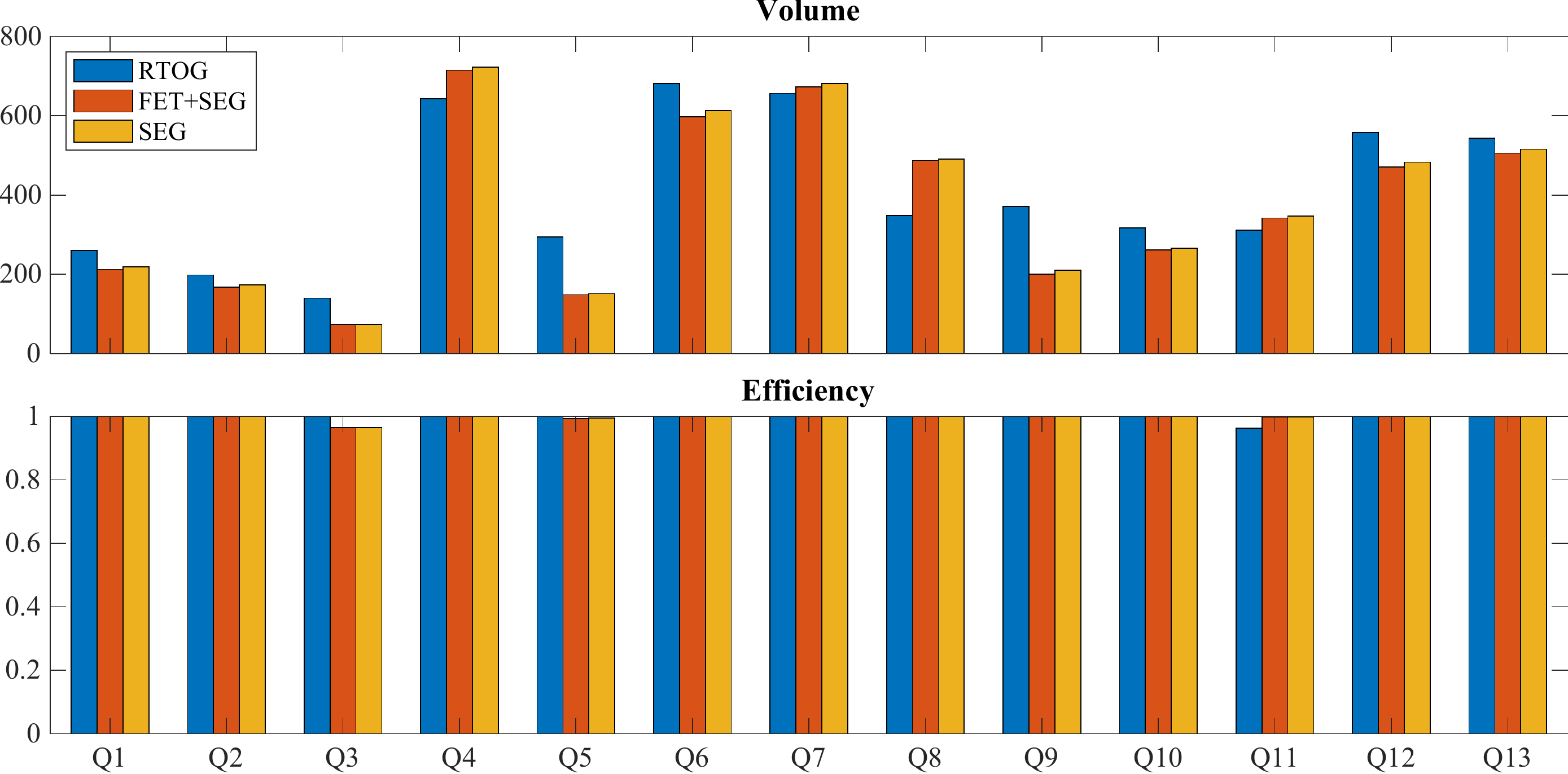}
  \caption{
    Comparison of RTOG CTV (blue) with Personalized CTV (1\% isosurface of predicted tumor cell density) from different parameter sources: \petseg{} (red), and \segonly{} (yellow).
    (A) Total irradiated volume.
    (B) Efficiency, as percentage of recurrent tumor core covered by the CTV.
    Personalized CTV from \petseg{}, and \segonly{} present reduced or similar irradiation volumes while maintaining efficiency comparable to RTOG CTV.
  }
    \label{f:addvoleff_t1c}
\end{figure}

\begin{figure}[!h]
  \centering
  \includegraphics[keepaspectratio,width=\linewidth]{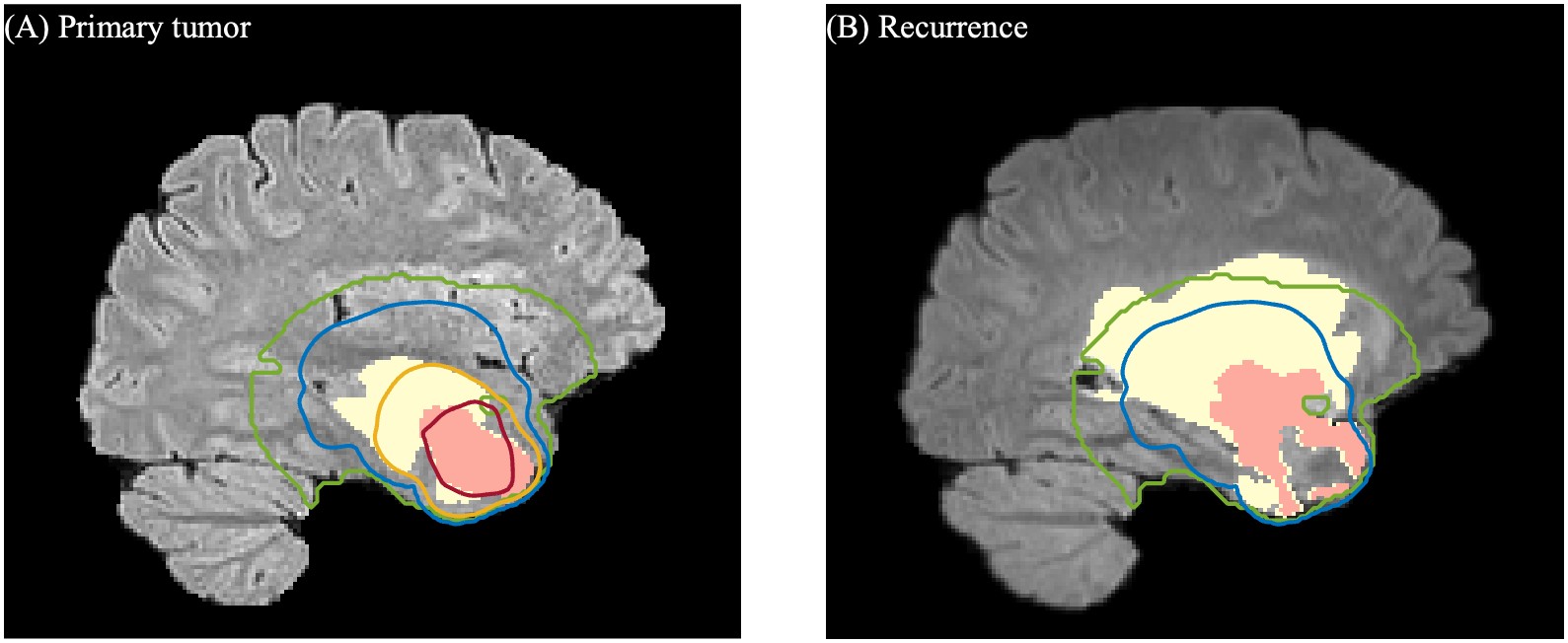}
  \caption{
    Comparison of  RTOG CTV (green) with Personalized CTV \petseg{} (blue) for patient Q13 over (A) pre-operative tumor and (B) recurrent tumor.
    Overlap of inferred segmentations $\ucf$ (yellow line) and $\uct$ (red line) on actual segmentations: T1Gd (beige fill) and FLAIR (pink fill); Margins for $\ctvrtog$ (green) and $\ctvp$ (blue).
   }
    \label{f:explainq3}
\end{figure}
% vol: 178 vs 73, eff 88 vs 66
Figure~\ref{f:voleff} shows that the personalized CTV from either \petseg{} or \segonly{}, with an efficiency of 66\% and volume of 73 mm\textsuperscript{3}, is less efficient and smaller compared to the RTOG CTV, which has an efficiency of 88\% and a volume of 178 mm\textsuperscript{3}.
We investigate this further in Fig.~\ref{f:explainq3}, where we show the sagittal view of (A) the pre-operative tumor and (B) the recurrent tumor.
The prediction based on pre-operative tumor is good, with DICE\textsuperscript{T1Gd} 0.76 and DICE\textsuperscript{FLAIR} 0.69.
Both the recurrent tumor core and the edema regions are larger than the corresponding pre-operative regions.
Quantitatively, the percentage of CTV that covers the recurrent tumor core is 44\% for RTOG CTV and 63\% for the personalized CTV from \petseg{}, which suggests that the personalized CTV is capturing the infiltration pathway, as it spares more normal tissue than RTOG CTV.

Several factors may contribute to the notably large size of the recurrent tumor.
Firstly, since the post-operative tumor core region is also larger than the pre-operative tumor core region, it is possible that the resection was not complete, and the residual tumor cell keeps growing.
Secondly, acute reactive changes after surgery or radiation can disrupt the fluid regulation in the brain, leading to edema \citep{ohmuraPeritumoralEdemaGliomas2023}.
While our study assumes the edema region as part of the tumor, indicating tumor infiltration \citep{ohmuraPeritumoralEdemaGliomas2023}, the differentiation between peritumoral edema and glioma invasion remain a significant challenge and is an active area of research \citep{ohmuraPeritumoralEdemaGliomas2023}.

%%%%%%%%%%%%%%%%%%%%%%%%%%%%%%
% \section{Additional Results Patient Dataset B1-B6}

\pgfplotstableread[col sep=comma]{tab_Q_compare.csv}\mydata
\begin{table*}[!t]
  \centering
  \pgfplotstabletypeset[
    string type,
    header=false,
    col sep=comma,
    every last row/.style={after row=\hline},
    every first row/.style={before row=\toprule},
    every head row/.style={output empty row},
    every nth row={2[+1]}{before row=\hline}
    ]\mydata
    \caption{Estimated parameters and metrics for Patients Q1-Q16, comparison between results using \segonly{} and \petseg{} data.}
    \label{t:newpatient}
  \end{table*}

%%%%%%%%%%%%%%%%%%%%%%%%%%%%%% 
\section{Estimation of characteristic parameters}
\label{ap:char}

We ignore the geometry of the brain and consider a spherically symmetric geometry: a homogenous sphere of radius R sufficiently larger than the brain (here we use $R=180$).
Taking $\muD$ = 1, $\muR$ = 1, and $P(\vb x) = 1$ then the PDE \eqref{eq:pde} reduces to the radially symmetric system $u(r,t)$, where $r\in [0, R]$, $t \in [0 ,1]$,
\begin{equation}
  \begin{cases}
    \pdv{u}{t} = \cD \frac{1}{r^2} \pdv{r} \left( r^2 \pdv{r} u \right)  +   \cR u (1-u) \\
    \pdv{u}{r} = 0 \quad \text{on } r = 0 \quad\text{and } r = R \\
    u = 0.1 \exp(-0.1 r)  \quad \text{at } t = 0 \\
  \end{cases}
  \label{eq:pdesph}
\end{equation}

The detailed procedure for estimating the characteristic parameters is as follows:
\begin{itemize}    
  \item Compute the centroid of the FLAIR segmentation $\segt$, denoted as $\vbx_0$, which is used as the initial guess for the tumor center.
  \item Find the radius $\rf_{\rm seg}$ such that the sphere with center $\vbx_0$ and radius $\rf_{\rm seg}$ encloses the FLAIR segmentation $\segf$:
  \begin{equation*}
    \rf_{\rm seg} = \max \{ r | r = \parallel\vbx - \vbx_0\parallel, \segf(\vbx) = 1 \}
  \end{equation*}
  Similarly find the radius $\rt_{\rm seg}$ for $\segt$.
  \item For $\ra = 0.1:0.1:1$ (matlab notation) and $L=10:5:90$, 
  solve \eqref{eq:pdesph}.
  Based on \jana{}, we assume $\ucf = 0.35$ and $\uct=0.6$.
  Record the radii corresponding to $\ucf$ and $\uct$ as $\rf_{\rm sph}$ and $\rt_{\rm sph}$ 
  \item find the combination $\ra$ and $L$ such that sum of the relative error 
  $|\rf_{\rm seg} - \rf_{\rm sph}|/\rf_{\rm seg}$ +  
  $|\rt_{\rm seg} - \rt_{\rm sph}|/\rt_{\rm seg}$
  is minimized.
  % We constrain $L>\rf_{\rm seg}$ as we want the normalized coordinate to be closer to the range $[0,1]$.
  \item This yields the patient specific characteristic parameters $\ra$, $L$. 
  We solve the PDE in the brain geometry with $\muD = \muR = 1$ to obtain the characteristic solution $\bufdm$.
\end{itemize}

We show the characteristic parameters in Table~\ref{t:charsyn} (synthetic cases S1-S8), Table~\ref{t:charjana} (patients P1-P8), and Table~\ref{t:charadd} (patients Q1-Q16).

\pgfplotstableread[col sep=comma]{tab_syn_char.csv}\mydata
\begin{table*}[!t]
  \centering
  \pgfplotstabletypeset[
    string type,
    header=false,
    col sep=comma,
    every last row/.style={after row=\hline},
    every first row/.style={after row=\hline,before row=\toprule},
    every head row/.style={output empty row},
    ]\mydata
    \caption{Characteristic parameter estimation from synthetic data. Datasets S1, S3, S5, and S8 use $\vbx_0$ = (164, 116, 99) while S2, S4, S6, and S7 use $\vbx_0$ = (65, 70, 99). For this synthetic data, the estimated values for $\muD$ and $\muR$ are near $1$, which is the targeted outcome.}
    \label{t:charsyn}
  \end{table*}

\pgfplotstableread[col sep=comma]{tab_jana_char.csv}\mydata
\begin{table*}[!t]
  \centering
  \pgfplotstabletypeset[
    string type,
    header=false,
    col sep=comma,
    every last row/.style={after row=\hline},
    every first row/.style={after row=\hline,before row=\toprule},
    % every head row/.style={before row=\hline, after row=\hline},
    every head row/.style={output empty row},
    ]\mydata
    \caption{Characteristic parameter estimation for Patients P1-P8. Parameters $D$, $\rho$, and $\tend$ are from \jana{} and used to compute $\muD$ and $\muR$.
    }
    \label{t:charjana}
\end{table*}

\pgfplotstableread[col sep=comma]{tab_char_Q.csv}\mydata
\begin{table*}[!t]
  \centering
  \pgfplotstabletypeset[
    string type,
    header=false,
    col sep=comma,
    every last row/.style={after row=\hline},
    every first row/.style={after row=\hline,before row=\toprule},
    every head row/.style={output empty row},
    ]\mydata
    \caption{Characteristic parameter estimation for Patients Q1-Q16.}
    \label{t:charadd}
  \end{table*}

%%%%%%%%%%%%%%%%%%%%%%%%%%%%%%
\section{Effect of different types of data loss}
\label{ap:datatype}

We perform a series of experiments using synthetic data S1-S8 to study the effect of different types of data in the fine-tuning stage.
As in Section~\ref{ss:loss}, let $u(\vb x, t)$ be the function represented by the PINN.
For the synthetic dataset, we have the ground truth cell density $u^{GT}$. Unlike the main text, here we do not perturb the $u^{GT}$ with noise.
As in Eq.\eqref{eq:totalloss}, the total loss for fine-tuning has three components: residual loss $\lres$, data loss $\ldat$, and regularization loss $\mathcal{L}_{\rm \beta}$ for the parameters $\beta \in \Theta$:
\begin{equation}
  \ltot = \lres + \ldat + \sum_{\beta \in \Theta} \mathcal{L}_{\rm \beta},
\end{equation}
where $\ldat$ depends on the type of data used for fine-tuning. 
We have studied the results using \petseg{} or \segonly{} only data, and we consider two additional types of data and their corresponding data loss functions:
\begin{itemize}
  \item ground truth cell density at residual collocation points (time goes from [0,1]), denoted as \utall{} in Table~\ref{t:datatype}. 
  \begin{equation*}
    \ldat = \frac{1}{\Nres}\sum_{i=1}^{\Nres} \left(\phi(\xres_i) u(\xres_i, t_i) - \phi(\xres_i)u^{GT}(\xres_i, t_i) \right)^2.
  \end{equation*}
  \item ground truth cell density at data collocation points (at time 1), 
  denoted as \utend{} in Table~\ref{t:datatype}.
  \begin{equation*}
    \ldat = \frac{1}{\Ndat}\sum_{i=1}^{\Ndat} \left(\phi(\xdat_i) u(\xdat_i, 1) - \phi(\xdat_i)u^{GT}(\xdat_i, 1) \right)^2.
  \end{equation*}
\end{itemize}
These two cases do not involve the imaging model, and the trainable parameters are $\{\muD, \muR, \icx\}$.

% These cases are ordered in terms of the intuitive difficulty of the data loss. 
In Table~\ref{t:datatype}, we compare the relative error of $\muD$ and $\muR$, which are the most important parameters for tumor growth prediction.
We see that using the cell density data at all times gives the smallest relative error.
Using the cell density data at time 1 gives the second smallest relative error.
Using the segmentation and FET data gives similar errors as using the segmentation data only, 
and both have larger errors than using the cell density data. Of course, the cell density data is not available from current imaging technologies.
% In the synthetic case, since the both the segmentation and PET are generated based on cell density.

\begin{table*}[h]
  \centering
  \begin{tabular}{l c c c c  c c c c}
      \hline
      & \multicolumn{4}{c}{$\muD$} & \multicolumn{4}{c}{$\muR$} \\
      \cmidrule(r{4pt}){2-5} \cmidrule(l{4pt}){6-9}
      case & \utall & \utend{} & \petseg & \segonly & \utall & \utend{} & \petseg & \segonly \\
      \hline
      median & 5.56 & 6 & 7.48 & 7.58 & 0.628 & 2.93 & 1.35 & 1.25 \\
      mean & 5.62 & 6.92 & 9.7 & 9.62 & 1.09 & 2.54 & 1.37 & 1.45 \\
      std & 4.74 & 3.99 & 8.86 & 8.45 & 1.2 & 1.53 & 1.07 & 0.823 \\
      \hline
  \end{tabular}
  \caption{Statiscis of absolute relative error (\%) for $\muD$ and $\muR$ in the synthetic dataset (S1-S8) using varied data losses for fine-tuning. \utall{} uses the ground truth cell density at residual collocation points over times in [0,1]. Meanwhile, \utend{} uses ground truth cell density at data collocation points at time 1. \petseg{} uses noisy segmentation and PET data. \segonly{} uses noisy segmentation data only.}
  \label{t:datatype}
\end{table*}

\section{Correlation of density with FET signals}
\label{ap:petcorr}

\begin{figure}[!h]
  \centering
  \includegraphics[keepaspectratio,width=\linewidth]{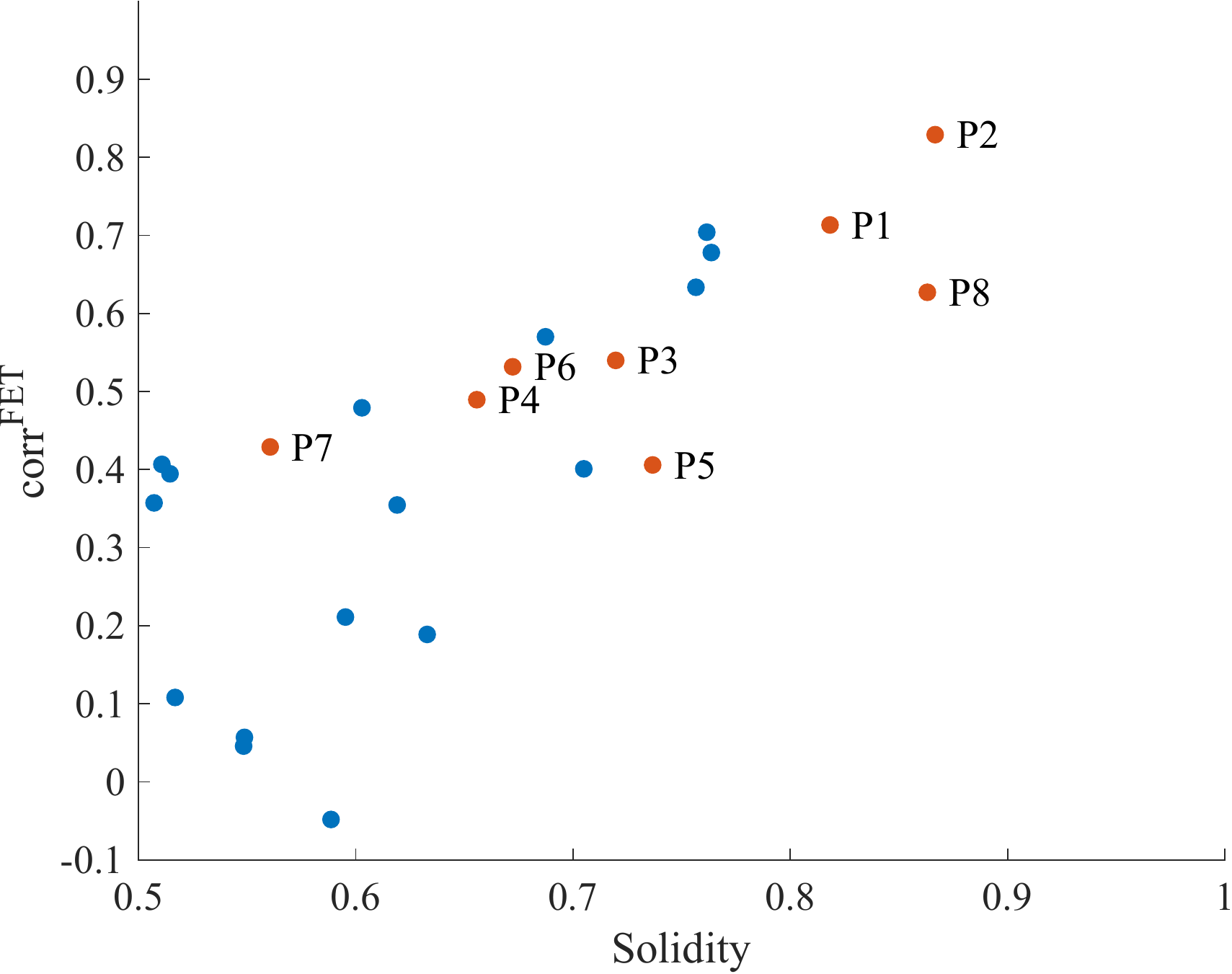}
  \caption{Correlation between FET signal and solidity of the FET region. The correlation is 0.757 with p-value 1.8e-5.
  Ring-like FET signals has solidity values lower than 1, indicative of a central void, and is related to lower correlation between predicted cell densities and FET signals.
  }
  \label{f:corvssol}
\end{figure}

We observed considerable variability in \corfet{}, as shown in Table~\ref{t:allpat}.
Visually, these signals manifest in distinct patterns: some patients (such as P5, P7 in Fig.~\ref{f:patsegpet}) exhibit a ring-like appearance encircling the tumor, while others (P1, P2, P8) display a more uniform, blob-like signal.
To quantify the apparent central signal void in some tumors, we calculated the 'solidity' of the FET region, defined as the ratio of the FET region's area to the area of its convex hull. 
This metric effectively differentiates between morphologies: ring-like signals yield solidity values significantly lower than 1, indicative of a central void, whereas blob-like signals approach a solidity of 1, reflecting a more homogenous signal distribution.
Figure~\ref{f:corvssol} demonstrates a notable positive correlation between \corfet{} and the solidity of the FET region, indicating that our initial assumption of a linear relationship between cellular density and FET uptake may not universally apply, particularly in cases exhibiting ring-like signal patterns.

Two factors might contribute to the ring-like FET signals. 
Firstly, tumor necrosis, a common feature in GBM, results in non-viable, metabolically inactive tissue, which inherently shows reduced FET uptake \citep{meyer18FFETPET2021,lafougereMolecularImagingGliomas2011}.
Secondly, the partial volume effect might also play a role that is not captured by our image model \citep{soretPartialVolumeEffectPET2007}
Since FET-PET spatial resolution is 4 mm, larger than the 1 mm resolution of MRI, 
signal averaging between active tumor regions and adjacent necrotic or non-tumor areas can lead to an underestimation of tracer uptake.
These findings suggest that a more nuanced model may be required to accurately interpret FET signals, especially in the presence of heterogeneous tumor morphologies. 
Refining our model to account for these variations presents a promising avenue for future research.

\section{Effect of Personalized characteristic parameters}
\label{ap:fixchar}

We use synthetic data to demonstrate the importance of using personalized characteristic parameters for pre-training in our method.
W consider an alternative approach that use the pre-trained network based on S7 (see Table~\ref{t:charsyn}) for fine-tuning.
The average length scale $\bL$ and ratio $\ra$ of all synthetic cases are $\bL = 37$ and $\ra = 14.8$, which are close to the values for S7 ($\bL = 35$ and $\ra = 14$). 
Therefore we can also consider the pre-trained model based on S7 as a pre-trained model with ``average'' characteristic parameters.

In Table.~\ref{t:fixchar},
we compare the relative error of estimated parameters and the performance metrics of two cases:
(1) cases S1, S2, S5, S6, which use personalized pre-training, as presented in the main text;
and (2) cases S1$^*$, S2$^*$, S5$^*$, S6$^*$, which use the pre-trained network based on S7 for fine-tuning.
We see that personalized pre-training recovers the parameters more accurately, and achieves better performance metrics, while using a fixed pre-trained model gives worse results.
Note that S2 and S5 also have the same initial location $\icx$ as S7, therefore there is no change in the geometry between pre-training and fine-tuning.
We believe the suboptimal performance is mainly because the tumors exists different scales (different size and profile, as described by $\bL$ and $\ra$), and the pre-trained model based on S7 can not capture the large variations in the tumor morphology.
This leads to challenge in fine-tuning, as the ``true'' solution might be far from the initial guess, and the training might get stuck in a local minimum.
Note that in this invesitagion, the pre-trained model is based on one set of characteristic parameters. 
Pre-training a single network for all patients with different $rD$, $rR$ and geometry will be a promising future direction.

\begin{table*}[h!]
  \centering
  \begin{tabular}{lccccccccc}
    \hline
    Case & $\muD$(\%) & $\muR$(\%) & m(\%) & A(\%) & $\ucf$ (\%)& $\uct$ (\%)& DICE\textsuperscript{FLAIR} & DICE\textsuperscript{T1Gd} & Corr\textsuperscript{FET} \\
    \hline
    S1 & -1.36 & -0.90 & -5.64 & -16.00 & 2.03 & 2.79 & 0.92 & 0.86 & 0.98 \\
    S1$^*$ & 163.00 & 10.60 & 20.00 & 49.00 & -1.37 & -16.70 & 0.56 & 0.36 & 0.91 \\
    \hline
    S2 & -3.74 & -0.90 & -3.03 & 12.00 & 8.95 & 5.77 & 0.93 & 0.90 & 0.98 \\
    S2$^*$ & 191.00 & 12.40 & 20.00 & 238.00 & 42.80 & 3.09 & 0.78 & 0.66 & 0.88 \\
    \hline
    S5 & 0.20 & 1.55 & -3.86 & -26.40 & 0.02 & 0.00 & 0.96 & 0.81 & 0.98 \\
    S5$^*$ & -51.00 & -12.30 & -2.00 & -56.40 & -0.01 & -4.17 & 0.51 & 0.38 & 0.87 \\
    \hline
    S6 & 4.52 & -2.39 & -0.13 & -30.30 & 0.00 & -3.81 & 0.94 & 0.87 & 0.98 \\
    S6$^*$ & -48.90 & -8.47 & -9.69 & -59.30 & -0.01 & 2.28 & 0.60 & 0.58 & 0.90 \\
    \hline
  \end{tabular}
  \caption{Percentage relative error of estimated parameters and the metrics, comparing effect of personalized pre-training (lableled S[1256]) and fixed pre-training based on pre-trained model using characteristic parameter of S7 (lableled S[1256]$^*$). Personalized characteristic parameters give better results.
  }
  \label{t:fixchar}
\end{table*}

\bibliographystyle{model2-names.bst}
\biboptions{authoryear}
\bibliography{pinnglioma.bib}

\end{document}